\documentclass{article}

    \PassOptionsToPackage{numbers, compress}{natbib}
\usepackage[preprint]{neurips_2025}

\usepackage[utf8]{inputenc} % allow utf-8 input
\usepackage[T1]{fontenc}    % use 8-bit T1 fonts
\usepackage{hyperref}       % hyperlinks
\usepackage{url}            % simple URL typesetting
\usepackage{booktabs}       % professional-quality tables
\usepackage{amsfonts}       % blackboard math symbols
\usepackage{nicefrac}       % compact symbols for 1/2, etc.
\usepackage{microtype}      % microtypography
\usepackage{xcolor}         % colors
\usepackage{amsmath}
\usepackage{amssymb}
\usepackage{mathtools}
\usepackage{amsthm}
\usepackage{multirow}
\usepackage{graphicx}
\usepackage{colortbl}
\usepackage{arydshln}
\usepackage{enumitem}
\usepackage{subfigure}
\usepackage{wrapfig}
\usepackage{algorithm}
\usepackage{algpseudocode}
\usepackage{subcaption}
\usepackage{float}
\usepackage{adjustbox}

\usepackage[capitalize,noabbrev]{cleveref}
\usepackage{bbding}
\usepackage{pifont}
\usepackage{minitoc}

\definecolor{myblue}{RGB}{224,247,250}
\definecolor{myblue2}{RGB}{186,230,251}

\definecolor{backblue}{RGB}{210, 230, 250}

\definecolor{backred}{RGB}{255, 223, 223}

\definecolor{backgreen}{RGB}{220,244,229}

\definecolor{back_deepblue}{RGB}{180, 210, 240}

\definecolor{back_deepred}{RGB}{255, 200, 200}

\definecolor{back_deepgreen}{RGB}{190, 230, 210}

\definecolor{mygray}{gray}{0.95}
\definecolor{greentable3}{rgb}{0,0.5,0}
\title{Beyond Examples: High-level Automated Reasoning Paradigm in In-Context Learning via MCTS}

\author{
    Jinyang Wu\textsuperscript{1}\thanks{\quad Equal contributions.\\} \quad
    \textbf{Mingkuan Feng}\textsuperscript{1}\footnotemark[1] \quad  
    \textbf{Shuai Zhang}\textsuperscript{1}\footnotemark[2]\quad
    \textbf{Feihu Che}\textsuperscript{\textbf{2}}\\
    \textbf{Zhengqi Wen}\textsuperscript{\textbf{2}}\quad
    \textbf{Chonghua Liao}\textsuperscript{\textbf{3}}\quad
    \textbf{Jianhua Tao}\textsuperscript{\textbf{1}}\textsuperscript{\textbf{2}}\footnotemark[2]\\
    \textsuperscript{1} Department of Automation, Tsinghua University \\
    \textsuperscript{2} Beijing National Research Center for Information Science and Technology \\
    \textsuperscript{3} Institution for Interdisciplinary Information Sciences, Tsinghua University \\
    \texttt{wu-jy23@mails.tsinghua.edu.cn}, zhang$\_$shuai@mail.tsinghua.edu.cn
}

\begin{document}
\doparttoc
\faketableofcontents

\maketitle

\begin{abstract}
In-context learning (ICL) enables large language models (LLMs) to perform downstream tasks through advanced prompting and high-quality demonstrations. However, traditional ICL paradigms encounter significant limitations in complex reasoning tasks, stemming primarily from their dependence on example quality and absence of explicit reasoning guidance. To address these challenges, we introduce HiAR-ICL, a \textbf{Hi}gh-level \textbf{A}utomated \textbf{R}easoning paradigm in \textbf{ICL} that shifts focus from specific examples to abstract reasoning patterns, thereby extending the conventional concept of ``context'' in ICL. Our approach begins by defining five atomic reasoning actions, upon which we employ Monte Carlo Tree Search to systematically construct high-level reasoning patterns. During inference, HiAR-ICL dynamically selects appropriate reasoning patterns based on problem attributes, providing explicit guidance for the model's reasoning process. Experiments demonstrate HiAR-ICL's effectiveness and efficiency: utilizing only 200 prior samples with Qwen2.5-7B-Instruct, our method achieves 80.6$\%$ accuracy on MATH and 62.5$\%$ on AMC, exceeding GPT-4o's 77.2$\%$ and 57.5$\%$. Our approach enhances performance across models of varying sizes while generalizing effectively across domains. Further analysis reveals that HiAR-ICL can also serve as a plug-and-play inference method compatible with post-training techniques like GRPO. Code and data are available at \href{https://github.com/jinyangwu/HiARICL}{https://github.com/jinyangwu/HiARICL}.
\end{abstract}

\section{Introduction}\label{sec1}
\noindent\textit{``Give a man a fish and you feed him for a day. Teach a man to fish and you feed him for a lifetime.''} 
\newline
\rightline{--- \textit{An old proverb}}

Large language models (LLMs) have demonstrated remarkable capabilities across diverse tasks and domains~\citep{zhao2023survey,achiam2023gpt,dubey2024llama}. Their proficiency in complex reasoning, particularly in mathematical tasks, has emerged as a critical benchmark for evaluating fundamental cognitive abilities~\citep{hao-etal-2023-reasoning,xi2024training,besta2025reasoning}. Mastering multi-step reasoning capabilities requires rigorous adherence to intricate rules and sophisticated problem-solving strategies, presenting significant challenges for existing LLMs~\citep{fu2023specializing,ahn-etal-2024-large}.

In-context learning (ICL) has emerged as a promising approach for enhancing LLMs' reasoning capabilities, distinguished by its simplicity and parameter-free nature~\citep{zhou-etal-2024-mystery}. Rooted in analogy-based learning~\citep{NEURIPS2020_1457c0d6}, ICL strategically employs curated demonstration examples to help models uncover latent patterns and generate sophisticated reasoning trajectories. Recent research has focused on improving ICL performance through advanced prompt engineering, including instruction optimization~\citep{wang-etal-2023-self-instruct} and demonstration selection~\citep{luo2024incontext}. A key advancement is Chain-of-Thought (CoT)~\citep{wei2022chain}, which uses prompts like ``Let's think step by step'' to facilitate more structured problem-solving~\citep{sprague2024cot}.

\begin{figure*}[t]
  \includegraphics[width=\textwidth]{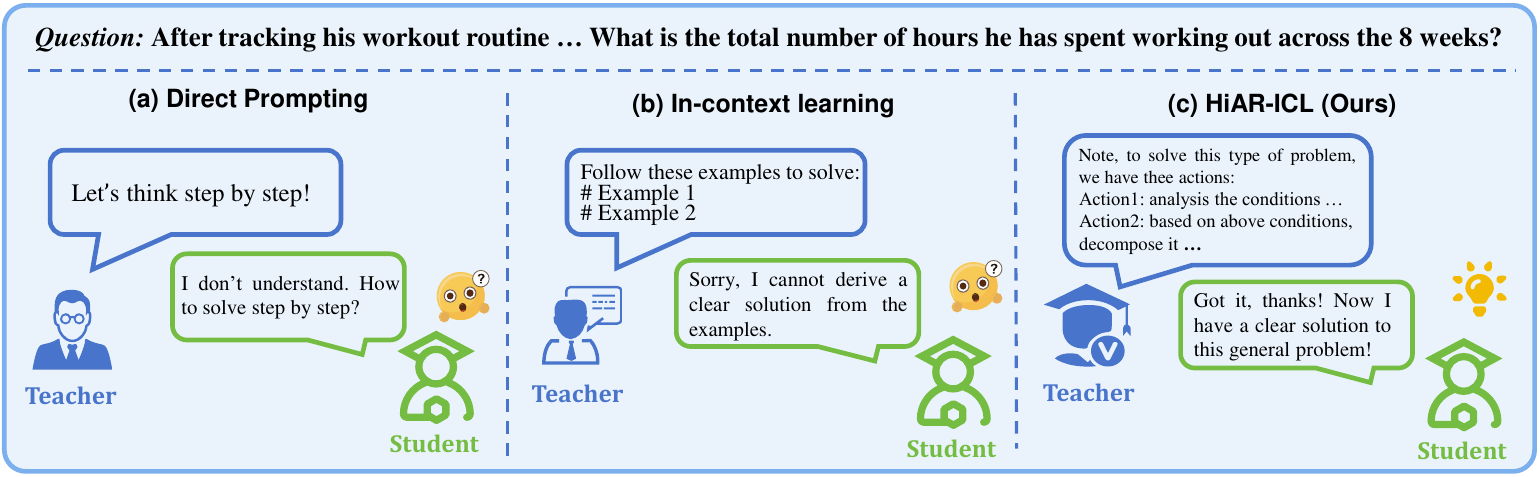}
  \caption{We compare HiAR-ICL with traditional zero-shot and few-shot in-context learning methods using a teacher-student paradigm. (a) Direct Prompting (Zero-shot CoT) provides a generic ``Let's think step by step'' instruction, which is insufficient for detailed step-by-step reasoning; (b) Few-shot In-Context Learning offers carefully selected examples but struggles with dissimilar problems; (c) Our method teaches high-level reasoning patterns, ensuring robust performance across various problems.}
  \label{Figure1}
\end{figure*}

Despite recent advances, current ICL paradigms face three critical limitations: (1) Empirical studies~\citep{wang-etal-2023-label,cui2024theoretical,shi-etal-2024-prompt,wangmixture} reveal that LLMs' reasoning performance is highly sensitive to demonstration characteristics such as quantity, ordering, and label distributions. Suboptimal examples can impede reasoning abilities, often requiring considerable human curation; (2) existing ICL methods lack explicit guidance signals, forcing LLMs to infer implicit cues through imitation~\citep{zhao-etal-2024-unveiling}; and (3) ICL's generalization capacity remains constrained. Reconstructing examples becomes necessary when encountering tasks with similar logical structures but different presentation formats~\citep{dong-etal-2024-survey}.

To address these challenges, we propose HiAR-ICL, a \textbf{Hi}gh-level \textbf{A}utomated \textbf{R}easoning paradigm in \textbf{ICL} through Monte Carlo Tree Search (MCTS). Our approach extends traditional ICL by redefining ``context'' from mere examples to higher-order cognitive reasoning patterns. This paradigm embodies the intuitive principle of ``teaching how to think, rather than what to think''. Figure \ref{Figure1} illustrates the core concepts of our method and traditional ICL approaches within a teacher-student paradigm.

Specifically, our method comprises two key steps: (1) MCTS-powered thought card construction, and (2) adaptive reasoning and verification. \underline{\textit{First}}, we define five atomic reasoning actions as the fundamental building blocks of chain-structured reasoning patterns (termed "thought cards" that serve as reference guidance during inference). Leveraging these actions and a small set of seed data, we apply MCTS to automatically derive high-level reasoning patterns, constructing multiple thought cards that capture diverse reasoning strategies. \underline{\textit{Second}}, during inference, we adaptively select thought cards based on problem characteristics and guide the model's reasoning process accordingly. This step incorporates verification through process reward modeling and self-consistency checks to validate the final solution. Extensive experiments demonstrate that HiAR-ICL significantly outperforms traditional ICL methods across diverse tasks and domains, even achieving comparable results to powerful closed-source models like GPT-4o with a 7B backbone. Our main contributions are:

\begin{itemize}[leftmargin=1.18em]
    \item \textbf{Novel ICL Insight}: We transcend traditional ICL by extending ``context'' from specific examples to abstract (higher-level) reasoning patterns, advancing the frontier of ICL research.
    \item \textbf{Automated Adaptive Reasoning Paradigm}: We propose an MCTS-powered framework that automatically generates diverse reasoning patterns and adaptively applies them based on problem characteristics, enabling robust reasoning performance and cross-domain generalization.
    \item \textbf{Remarkable Performance and Efficiency}: HiAR-ICL achieves 80.6$\%$ accuracy on MATH and 62.5$\%$ on AMC with Qwen2.5-7B-Instruct, surpassing GPT-4o (77.2$\%$ and 57.5$\%$), while reducing time cost by approximately 10$\times$ compared to leading test-time inference methods.
\end{itemize}

\begin{figure*}[ht!]
\vskip 0.02 in
\includegraphics[width=0.98\textwidth]{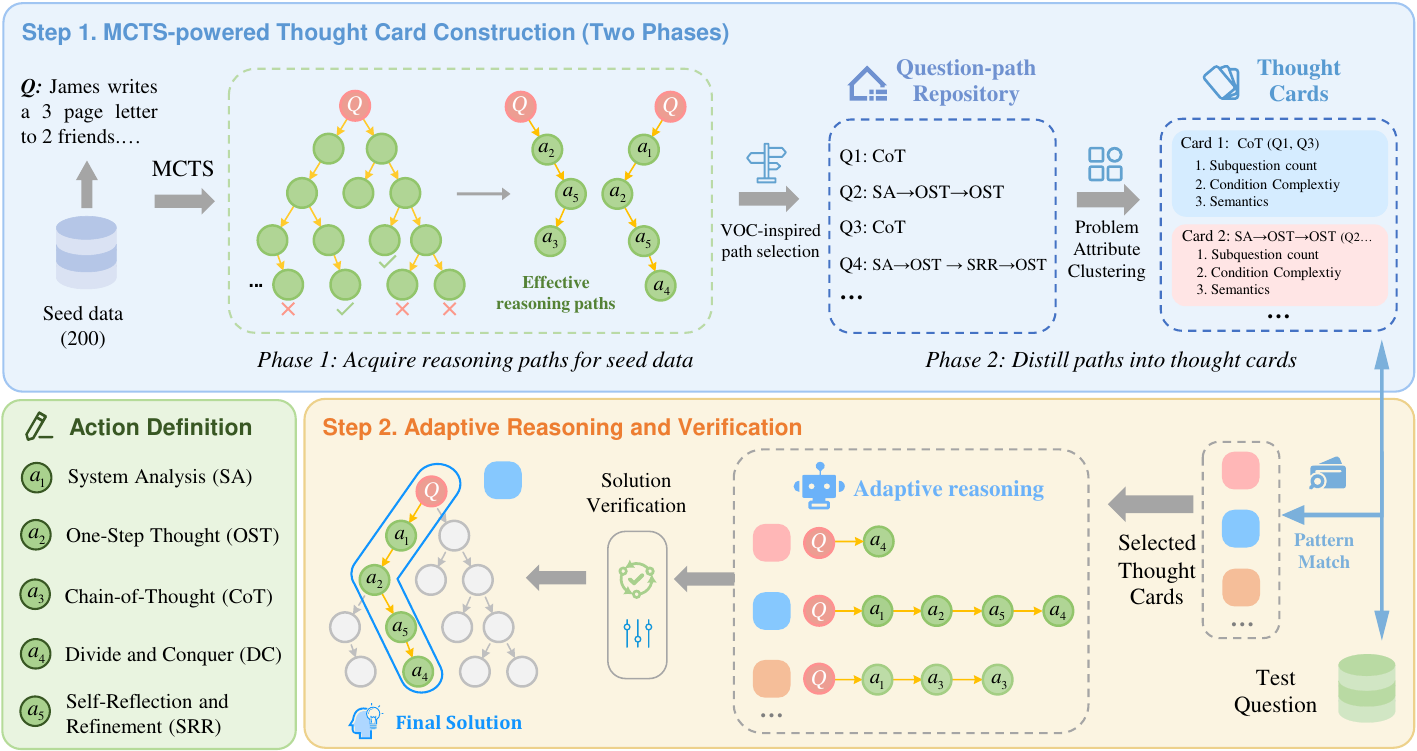}
\caption{Flowchart of our method HiAR-ICL. This framework consists of two main parts: (1) MCTS-Powered Thought Card Construction; and (2) Adaptive Reasoning and Verification.}
\label{Figure2}
\vskip -0.1 in
\end{figure*}

\section{Methodology}\label{sec4}
\textbf{Overview of HiAR-ICL} This section introduces HiAR-ICL in detail. As shown in Figure \ref{Figure2}, our approach consists of two main components:
\begin{itemize}[leftmargin=1.0em]
    \item \textit{MCTS-powered Thought Card Construction}: Leverage MCTS to systematically construct high-level thought cards, which effectively guides subsequent problem-solving.
    \item \textit{Adaptive Reasoning and Verification}: Dynamically select and execute optimal reasoning patterns based on the problem's cognitive complexity, followed by solution verification.
\end{itemize}

\subsection{MCTS-powered Thought Card Construction}\label{3.2}
Understanding human complex reasoning is crucial for modeling cognitive processes~\citep{Jaffe23}. Existing studies distinguish between two cognitive systems: System 1 and System 2~\citep{Kahneman2011,da2023system}. While ``System 1'' represents fast, intuitive, yet error-prone thinking, ``System 2'' involves slow, deliberative thinking with superior performance. With the emergence of advanced models like OpenAI's o1, developing efficient ``System 2'' approaches to emulate human cognitive processes has gained significant research attention~\citep{qin2024o1,sun2024beats}. Inspired by this, we introduce five atomic human-like reasoning actions to bridge the gap between model reasoning and human cognition: \textit{System Analysis (SA, $a_{1}$)}, \textit{One-Step Thought (OST, $a_{2}$)}, \textit{Chain-of-Thought (CoT, $a_{3}$)}, \textit{Divide and Conquer (DC, $a_{4}$)}, \textit{Self-Reflection and Refinement (SRR, $a_{5}$)}. Detailed descriptions are provided in Appendix \ref{C.1}.

Following the action definition, we introduce ``\textit{thought cards}'', which serve as structured reasoning templates to guide subsequent inference in Section \ref{3.3}. Using a small seed dataset, we derive reasoning paths (Phase 1) and distill them into multiple thought cards (Phase 2). These cards serve as prior insights for testing stage, facilitating efficient adaptive reasoning.

\textbf{Phase 1: Acquire reasoning paths for seed data}
$\;$ As shown in Figure \ref{Figure2}, we employ MCTS to iteratively optimize the solution search process, generating high-quality reasoning paths for the seed dataset. This design combines MCTS's iterative nature with LLMs' inherent reasoning capabilities, leading to enhanced search outcomes~\citep{NEURIPS2021_d5eca8dc,zhou2024language}. Following previous studies~\citep{wang2024litesearch,qi2024mutual}, we formulate each complex reasoning question $q$ as a tree search problem, where $q$ represents the root node and subsequent nodes denote reasoning steps (actions and corresponding outcomes) generated by a policy LLM $\pi_{\theta}$. We define the state $S_{t-1}$ as the trajectory $q, s_{1},...,s_{t-1}$, where $S_{0}=q$. The next step is sampled as $s_{t}\sim \pi_{\theta}(S_{t-1})$. To guide tree expansion, we define $Q(s)$ as the reward value for node $s$. Initially, all unexplored nodes are assigned $Q(s_{i})=0$. They are updated using a weighted average between the parent's current value and its child node's value: 
\begin{equation}
    Q(p) \leftarrow (1-\alpha)Q(p)+\alpha Q(s) \label{eq1}
\end{equation} 
where $\alpha$ is a discount factor for future rewards. For terminal nodes, following prior work~\citep{qi2024mutual}, we adopt the likelihood (confidence) of self-consistency majority voting as the reward value, enabling supervision-free generalization. Specifically, this step comprises four MCTS operations:

\textit{(1) Selection}. This operation identifies optimal nodes for expansion. Starting from the root node, a child node is chosen at each tree level until reaching a leaf node, defined as achieving maximum tree depth or arriving at an answer here. To balance the exploration and exploitation, we employ the well-known Upper Confidence Bounds applied to Trees (UCT)~\citep{11871842_29} for node selection:
\begin{equation}
    UCT(s) = Q(s) + w\sqrt{\frac{\ln N(p) }{N(s)} }  \label{eq2}
\end{equation}
where $Q(s)$ is the reward value for node $s$, $N(s)$ is the visit count, $p$ is the parent node, and $w$ is the exploration weight. The node with the highest UCT value is selected for subsequent phases, balancing exploration and exploitation.

\textit{(2) Expansion}. The selected node $s$ is expanded by sampling $n$ actions from $\pi_{\theta}$ and generating corresponding reasoning outcomes. These $n$ child nodes are then added to the tree.

\textit{(3) Simulation}. Starting from the selected node, we iteratively sample and expand nodes until reaching a terminal state (maximum depth or final answer node). To enhance efficiency, we implement an early termination strategy based on self-consistency~\citep{wang2023selfconsistency}. This strategy exploits the observation that repeatedly sampled actions at the same state likely indicate successful task completion. If the model's consistency score exceeds a threshold $c$, i.e., $SC(s) > c$, the simulation terminates early.

\textit{(4) Backpropagation}. Upon simulation completion, node information is updated along the simulation path $s_{0},...s_{d}$. Visit counts are incremented ($N(s) \leftarrow N(s) + 1$), and node value $Q(s)$ is propagated backward to its parent node $p$ using Equation \ref{eq1}.

\textbf{Phase 2: Distill paths into thought cards}
% \vspace{0.22em}
$\;$ After executing MCTS, we obtain a tree structure for each seed dataset question, yielding multiple valid reasoning paths. To identify the optimal reasoning trajectory, we draw inspiration from the concept of Value of Computation (VOC)~\citep{RUSSELL1991361}. VOC posits that intelligent systems should optimize the trade-off between computational benefits and costs. Accordingly, we propose a novel VOC-inspired selection metric:
\begin{equation}
    Score(q,p_{q}) = k\cdot R(p_{q}|q) - (1-k)\cdot C(p_{q}) \label{eq3}
\end{equation}
where $q$ is the task input, $p_{q}$ denotes a candidate reasoning path, and $k$ balances benefits against computational costs. Here, $R(p_{q}|x)$ denotes the path's final reward (defined as the leaf node's Q-value), while $C(p_{q})$ is the reasoning cost (defined as the number of actions in the sequence). 

For each question in the seed dataset, we select the optimal path $p_{best}$ with the highest $Score(q,p_{q})$ to build a \textit{Question-path repository} $D$ with one-to-one mappings. Inspired by metareasoning principles~\citep{RUSSELL1991361,de2024rational}, which advocate for adaptive reasoning strategies, we distill these question-path pairs into abstract thought cards $\mathbb{C}$ that represent high-level reasoning patterns abstracted from similar problems. This transformation is guided by three cognitive complexity indicators~\citep{lee2000problem}: (1) \textit{Subquestion Count (SC)}: The number of decomposed subproblems required to solve question $q$; (2) \textit{Problem Condition Complexity (PCC)}: The number of prior known conditions in $q$; and (3) \textit{Problem Semantics (PS)}: The semantic representation derived from the $q$'s linguistic features. Finally, each card contains a high-level thought template (e.g., $SA \to OST \to CoT$), along with the average SC, PCC, and PS metrics of questions sharing this template. In this paper, we utilize PCC as our default metric, with SC and PS serving as optional criteria. More complex designs are left for future work.

\subsection{Adaptive Reasoning and Verification}\label{3.3}
\textbf{Adaptive Reasoning}
\; During inference, for a test question $q_t$, we compute its SC, PCC, and PS metrics using Llama3-8B-Instruct. We then perform nearest neighbor matching~\citep{6809191} against pre-constructed thought cards $\mathbb{C}$ to identify the most relevant cards that best align with the question's attributes. These chosen cards serve as reference templates for $q_t$ in the final solution generation and verification. This selection process is formalized as:
\begin{equation}    
    NN_5(q_t, \mathbb{C}) = \mathop{\arg\min}_{\mathbb{C}_{q_t} \subseteq \mathbb{C}, |\mathbb{C}_{q_t}| = 5}  {\textstyle \sum_{c \in \mathbb{C}_{q_t}}}d(q_t, c) \label{eq33}
\end{equation}
where $NN_5(q_t, \mathbb{C})\subseteq \mathbb{C}$ denotes the five most relevant thought cards, determined by the distance function $d$. We employ distinct distance metrics for each attribute: absolute difference (e.g., $|PCC{q_t} - PCC{c}|$) for SC and PCC, and cosine similarity between the semantic representations of $q_t$ and each card's associated questions. These selected cards serve as high-level reasoning templates. Following these templates, the model systematically breaks down the reasoning process into sequential steps, with each step corresponding to a specific atomic action.

\textbf{Verification}
\; To identify the most accurate solution among candidates~\citep{uesato2022solving,qi2024mutual,zhang2024rest}, we introduce a simple yet effective two-stage verification method. We first apply process-supervision scoring to evaluate each reasoning path. The top-3 highest-scoring paths then undergo self-consistency checks to determine the final solution. Our experiments confirm that even straightforward self-consistency checks can effectively identify precise reasoning chains without additional supervision.

\section{Experimental Settings}
\textbf{Datasets}
$\,$ We conduct experiments across multiple reasoning tasks. We use 200 randomly sampled training instances as seed data for thought card construction (Section \ref{3.2}), with evaluation on corresponding test sets: (1) arithmetic reasoning: GSM8K$_{1319}$~\citep{cobbe2021training} and SVAMP$_{300}$~\citep{Patel2021AreNM}; (2) mathematical reasoning: MATH$_{500}$~\citep{hendrycks2021measuring} and AMC$_{40}$~\citep{li2024numinamath}; (3) commonsense reasoning: StrategyQA$_{687}$~\citep{geva2021did}; (4) PHD-level Scientific Reasoning: GPQA$_{Diamond198}$~\citep{rein2024gpqa}. While Table \ref{table1} presents results across all datasets, our subsequent analysis mainly focus on MATH, GSM8K, SVAMP, and StrategyQA to enable direct comparisons with baselines that lack results on the challenging AMC and GPQA.

\textbf{Baselines}
We evaluate HiAR-ICL against three strong baseline categories: (1) traditional example-based ICL methods, including zero-shot CoT~\citep{kojima2022large}, few-shot CoT~\citep{wei2022chain}, and CoT+SC~\citep{wang2023selfconsistency}; (2) tree-based methods, including ToT~\citep{NEURIPS2023_271db992}, RAP~\citep{hao-etal-2023-reasoning}, ReST-MCTS*~\citep{zhang2024rest}, LLaMA-Berry~\citep{zhang2024llama}, and rStar~\citep{qi2024mutual}. (3) leading LLMs, including Llama3.1-405B~\citep{meta_llama_2023}, and GPT-4o~\citep{gpt4o}, and Claude-3.5~\citep{claude}. We provide more baseline comparisons like the advanced Reinforced ICL~\citep{agarwal2024manyshot} in Appendix \ref{E.1}.

\textbf{Evaluation Metrics}
$\,$ We evaluate our approach using two metrics: \textit{accuracy}, based on the strict matching between the model's final answer and ground truth, and \textit{average time cost per sample} to assess computational efficiency compared to existing leading search-based approaches.

\textbf{Implementation Details} $\,$ 
We use the vLLM framework\footnote{\url{https://github.com/vllm-project/vllm}} with parameters: temperature 0.8, top-p 0.9, top-k 40, repetition penalty 1.0, and max tokens 1024. For MCTS implementation, we set the maximum tree depth $d_{max}$ to 5, exploration weight $w$ to 2.0, and prediction terminal threshold $c$ to 0.90. In the VOC-based optimal path selection (Sec.~\ref{3.2}), we use a balance factor $k$ of 0.95. All experiments are conducted on a Ubuntu 22.04 machine equipped with NVIDIA A100-80GB GPUs.

\begin{table*}[ht!]
\centering
\caption{Accuracy ($\%$) of of different LMs with HiAR-ICL across six benchmarks. The best results in each box are highlighted in \textbf{bold}. All models are instruct versions.}
\label{table1}
\resizebox{1.0\linewidth}{!}{
\begin{tabular}{clccccccl}
\toprule
\multirow{2}{*}{\textbf{Model}}  & \multirow{2}{*}{\textbf{Method}} & \multicolumn{2}{c}{\textbf{Mathematics}} & \multicolumn{2}{c}{\textbf{Arithmetic}} & \textbf{Science} & \textbf{Commonsense} & \multirow{2}{*}{\textbf{Avg.}} \\
\cmidrule(lr){3-4} \cmidrule(lr){5-6} \cmidrule(lr){7-7} \cmidrule(lr){8-8}
 &  & MATH & AMC & GSM8K & SVAMP & GPQA & StrategyQA & \\
\midrule
\multirow{4}{*}{GPT-4o~\citep{gpt4o}} & Zero-shot CoT  & 70.2 & 42.5 & 91.0 & 92.0  & 53.6 & 74.1  & 70.5 \\
                                      & Few-shot CoT   & 76.6 & 52.5 & 94.3 & 93.0  & 54.0 & 72.5  & 73.8 \\
                                      & CoT+SC@4       & 77.2  & 57.5 & 94.3 & 94.0  & 54.5 & 79.5 & 76.2 \\
                                      & \cellcolor{mygray}Ours    & \cellcolor{mygray}\textbf{84.8} & \cellcolor{mygray}\textbf{62.5} & \cellcolor{mygray}\textbf{96.0} & \cellcolor{mygray}\textbf{94.7} & 
                                      \cellcolor{mygray}\textbf{58.1} &
                                      \cellcolor{mygray}\textbf{82.2} &  \cellcolor{mygray}\textbf{79.7} \\
\midrule
\multirow{4}{*}{Qwen2.5-14B~\citep{qwen25}} & Zero-shot CoT  & 69.8 & 42.5 & 92.4 & 91.6  & 23.3 & 62.8  & 63.7 \\
                                      & Few-shot CoT   & 80.0 & 45.0 & 94.8 & 91.3  & 28.3 & 53.1  & 65.4 \\
                                      & CoT+SC@4       & 76.2 & 60.0 & 94.0 & 91.0  & 32.3 & 69.7 & 70.5 \\
                                      & \cellcolor{mygray}Ours    & \cellcolor{mygray}\textbf{81.4} & \cellcolor{mygray}\textbf{65.0} & \cellcolor{mygray}\textbf{95.8} & \cellcolor{mygray}\textbf{93.7}  & \cellcolor{mygray}\textbf{44.4} &
                                      \cellcolor{mygray}\textbf{77.3}  &  \cellcolor{mygray}\textbf{76.3} \\
\midrule
\multirow{4}{*}{Qwen2.5-7B~\citep{qwen25}}  & Zero-shot CoT  & 64.8 & 32.5 & 86.2  & 91.3 & 24.7 & 52.8 & 58.7 \\
                                      & Few-shot CoT   & 68.6 & 47.5 & 91.6 & 92.3  & 34.3 & 67.6 & 67.0 \\
                                      & CoT+SC@4       & 76.4 & 50.0 & 92.0 & 92.3  & 34.3 & 73.2 & 69.7 \\
                                      & \cellcolor{mygray}Ours    & \cellcolor{mygray}\textbf{80.6} & \cellcolor{mygray}\textbf{62.5} & \cellcolor{mygray}\textbf{93.7} & \cellcolor{mygray}\textbf{93.0}  & \cellcolor{mygray}\textbf{43.4}  &
                                      \cellcolor{mygray}\textbf{76.0} &  \cellcolor{mygray}\textbf{74.9} \\
\midrule
\multirow{4}{*}{Qwen2-7B~\citep{yang2024qwen2}}    & Zero-shot CoT  & 36.9 & 12.5 & 76.6 & 85.2  & 19.7 & 55.3 & 47.7\\
                                      & Few-shot CoT   & 52.9 & 17.5 & 85.7 & 87.3  & 19.7 & 62.3 & 54.2 \\
                                      & CoT+SC@4       & 55.6 & 20.0 & 87.7 & 90.3  & 26.7 & 65.5 & 57.6 \\
                                      & \cellcolor{mygray}Ours    & \cellcolor{mygray}\textbf{66.8} & \cellcolor{mygray}\textbf{30.0} & \cellcolor{mygray}\textbf{91.8} & \cellcolor{mygray}\textbf{92.7}  & \cellcolor{mygray}\textbf{40.4} &
                                      \cellcolor{mygray}\textbf{72.0}  &  \cellcolor{mygray}\textbf{65.6} \\
\midrule
\multirow{4}{*}{Yi-1.5-6B~\citep{young2024yi}}       & Zero-shot CoT  & 30.4 & 15.0 & 76.4  & 64.4 & 14.2 & 46.2 &  41.1 \\
                                      & Few-shot CoT   & 40.5 & 10.0 & 78.9 & 81.3  & 17.7 & 61.1 & 48.3 \\
                                      & CoT+SC@4       & 42.2 & 12.5 & 79.4 & 87.6  & 19.2  & 65.2 & 51.0 \\
                                      & \cellcolor{mygray}Ours    & \cellcolor{mygray}\textbf{57.4} & \cellcolor{mygray}\textbf{20.0} & \cellcolor{mygray}\textbf{86.4} & \cellcolor{mygray}\textbf{91.3}  & \cellcolor{mygray}\textbf{31.8} &
                                      \cellcolor{mygray}\textbf{70.3}  &  \cellcolor{mygray}\textbf{59.6} \\
\midrule
\multirow{4}{*}{Llama3-8B~\citep{dubey2024llama}}  & Zero-shot CoT  & 5.8 & 5.0 & 68.3 & 70.9 & 11.1 & 57.2 & 36.3 \\
                                      & Few-shot CoT   & 17.8 & 7.5 & 74.5 & 81.0 & 22.7 & 68.4 & 45.3 \\
                                      & CoT+SC@4       & 28.8 & 15.0 & 80.6 & 88.0  & 24.7 & 66.8 & 50.7 \\
                                      & \cellcolor{mygray}Ours    & \cellcolor{mygray}\textbf{46.6} & \cellcolor{mygray}\textbf{30.0} & \cellcolor{mygray}\textbf{89.6} & \cellcolor{mygray}\textbf{92.7}  & \cellcolor{mygray}\textbf{39.4}  & \cellcolor{mygray}\textbf{73.4} &  \cellcolor{mygray}\textbf{62.0} \\
\midrule
\multirow{4}{*}{Llama3.1-8B~\citep{meta_llama_2023}}& Zero-shot CoT  & 18.0 & 10.0 & 61.5 & 69.3   & 27.2  & 52.4 & 39.7 \\
                                      & Few-shot CoT   & 47.2 & 12.5 & 76.6 & 82.0  & 23.3 & 63.6 & 50.8 \\
                                      & CoT+SC@4       & 44.2 & 15.0 & 80.5 & 85.6  & 29.8 & 69.8 & 54.1 \\
                                      & \cellcolor{mygray}Ours    & \cellcolor{mygray}\textbf{58.0} & \cellcolor{mygray}\textbf{32.5} & \cellcolor{mygray}\textbf{90.7} & \cellcolor{mygray}\textbf{93.0} &
                                      \cellcolor{mygray}\textbf{40.4} &\cellcolor{mygray}\textbf{75.7} & \cellcolor{mygray}\textbf{65.1} \\

\bottomrule
\end{tabular}}
\vskip -0.1in
\end{table*}

\section{Results and Discussion} 
This section validates the effectiveness of HiAR-ICL from four aspects: \textcolor{black}{\hyperlink{sec4.1}{4.1 performance}}, \hyperlink{sec4.2}{4.2 computational efficiency}, \hyperlink{sec4.3}{4.3 out-of-distribution generalization}, \hyperlink{sec4.5}{4.4 weak-to-strong generalization}, and \hyperlink{sec4.4}{4.5 ablation studies and analysis}.

\hypertarget{sec4.1}{\subsection{Performance Comparison with Baselines}\label{sec4.1}}
\textbf{Results on Diverse Reasoning Benchmarks}
$\;$ Table \ref{table1} presents the performance of HiAR-ICL across four mainstream reasoning benchmarks. We have three key findings: 

$\diamondsuit$ HiAR-ICL consistently outperforms traditional ICL methods across models. For example, Llama3-8B-Instruct's accuracy on MATH improved from 17.8$\%$ (few-shot CoT) to 46.6$\%$ (HiAR-ICL), a 2.6$\times$ improvement. Similarly, on AMC, performance increased from 7.5$\%$ (few-shot CoT) to 30$\%$ (HiAR-ICL), demonstrating the substantial potential of our approach.

$\diamondsuit$  The consistent performance gains across diverse reasoning tasks of different domains and varying difficulty levels further validate the generalizability of our high-level reasoning patterns. HiAR-ICL demonstrates robust improvements in arithmetic reasoning (GSM8K, SVAMP), mathematical reasoning (MATH, AMC), commonsense reasoning (StrategyQA), and PhD-level scientific knowledge reasoning (GPQA$_{Diamond}$). This cross-task effectiveness suggests that the high-level reasoning patterns transcend specific domain boundaries and difficulty levels.

\begin{wraptable}{r}{0.52\linewidth}
\vspace{-12.5pt}
\centering
\caption{Comparison with tree-based methods. The best results are highlighted in \textbf{bold}, with baseline results sourced from the original paper when accessible. All models are instruct versions.}
% \vskip 0.1in
\label{table2}
\resizebox{0.99\linewidth}{!}{
\begin{tabular}{clccc}
\toprule
\textbf{Model} & \textbf{Method} & \textbf{MATH} & \textbf{GSM8K} & \textbf{StrategyQA}\\
\midrule
\multirow{4}{*}{Qwen2-7B} & ToT & 53.3 & 79.0 & 66.7 \\
                          & RAP & 51.6 & 82.1 & 67.3 \\
                          & ReST-MCTS* & 52.4 & 82.3 & 64.9\\
                          & \cellcolor{mygray}Ours & \cellcolor{mygray}\textbf{66.8} & \cellcolor{mygray}\textbf{91.8} & \cellcolor{mygray}\textbf{80.9} \\
\midrule
\multirow{6}{*}{Llama3-8B} & ToT & 13.6 & 69.0 & 60.4 \\
                           & RAP & 18.8 & 80.5 & 68.7\\
                           & ReST-MCTS* & 34.2 & 75.5 & 65.0\\
                           & LLaMA-Berry & 39.6 & 88.1 & -\\
                           & rStar & 42.9 & \textbf{91.1} & 71.5\\
                           & \cellcolor{mygray}Ours & \cellcolor{mygray}\textbf{46.6} & \cellcolor{mygray}89.6 & \cellcolor{mygray}\textbf{73.4}\\
\bottomrule
\end{tabular}}
\vspace{-5pt}
\end{wraptable}

$\diamondsuit$ Our approach yields substantial improvements on small models, with Qwen2-7B-Instruct increasing from 52.9$\%$ to 66.8$\%$ and Yi-1.5-6B-chat from 40.5$\%$ to 57.4$\%$ on MATH. Similar enhancements are observed on PhD-level GPQA$_{Diamond}$ and olympiad-level AMC. These results highlight HiAR-ICL's potential for boosting the reasoning capabilities of small models.

\begin{figure}[t]
	\centering
	\begin{minipage}{0.38\textwidth}
		\centering
        % \vskip -0.2in
		\includegraphics[width=0.92\linewidth]{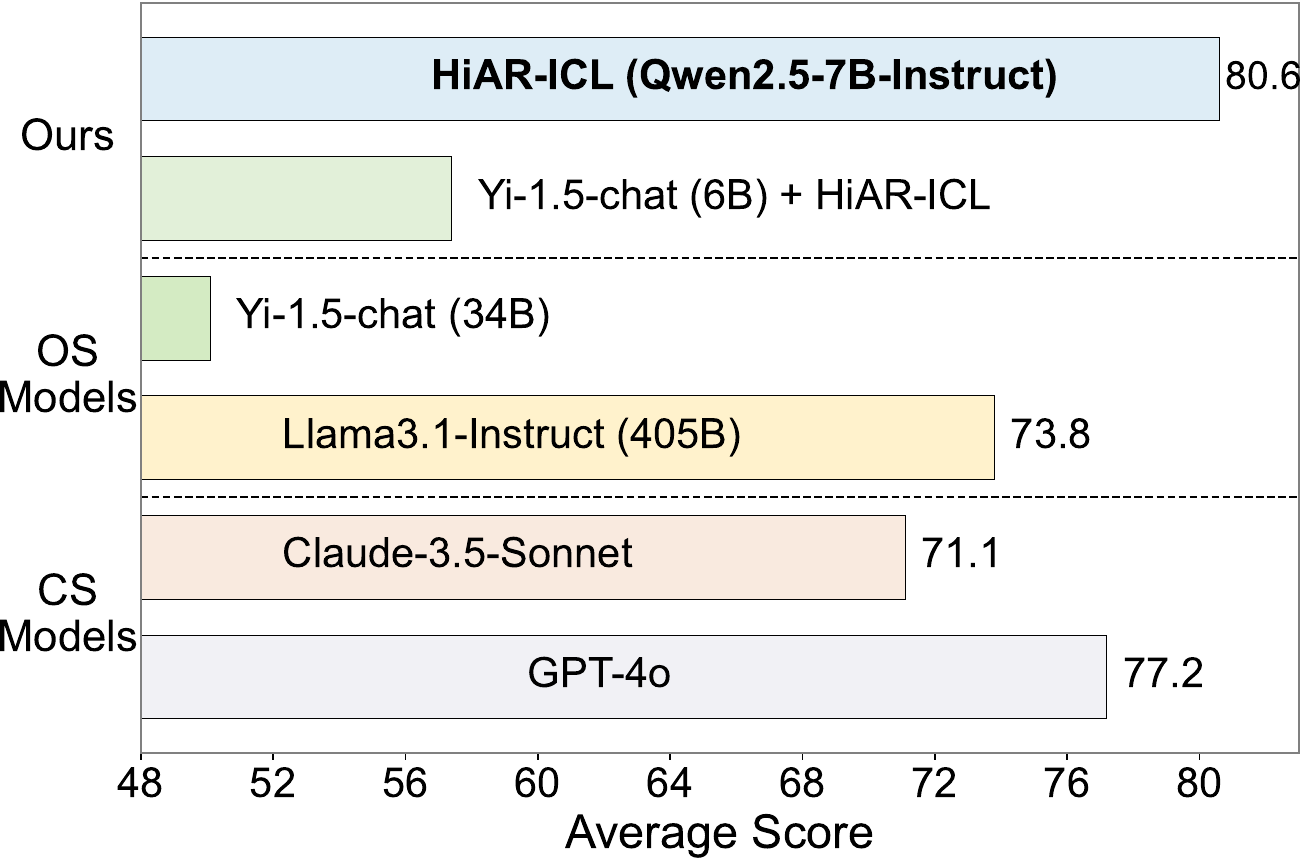} 
        % \vskip 0.05in
		\captionof{figure}{Intuitive Comparison on MATH. `OS' and `CS' denote open-source and closed-source models.}
		\label{Figure3}
	\end{minipage}
        \hfill
	\begin{minipage}{0.60\textwidth}
		\centering
        % \vskip -0.20in
		\includegraphics[width=0.98\linewidth]{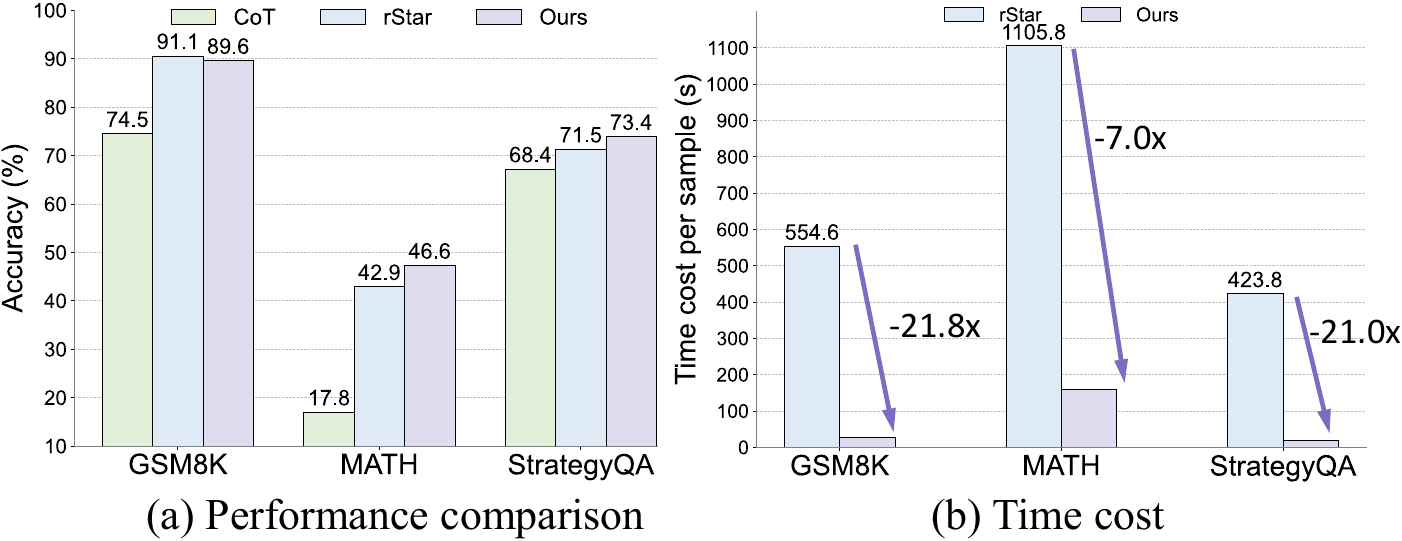} 
        \vskip 0.01in
		\captionof{figure}{Comparison with leading tree-based method, rStar. (a) Performance Comparison. (b) Total Time Cost Per Sample (the ``end-to-end'' time cost per sample).}
		\label{Figure4}
	\end{minipage}
% \vspace{0.1em}
\vskip -0.15in
\end{figure}

\textbf{Comparison with Powerful LLMs}
$\,$ We further investigate HiAR-ICL's performance upper bound when integrated with smaller models compared to larger models. As shown in Figure \ref{Figure3}, the 7B Qwen2.5 model achieves 80.6\% accuracy on the challenging MATH benchmark, outperforming few-shot GPT-4o (77.2\%). Similarly, Yi-1.5-6B+HiAR-ICL surpasses Yi-1.5-34B when employing our reasoning paradigm. This indicates that HiAR-ICL-empowered small models have the potential to exceed much larger models. In other words, while traditional ICL provides specific examples, our approach offers high-level problem-solving thought patterns that equip smaller models with problem decomposition capabilities typically found in larger models, thereby enabling superior performance. Detailed results are provided in Appendix Table \ref{tables30}.

\begin{figure*}[ht!]
\vskip 0.12 in
\centerline{\includegraphics[width=\linewidth]{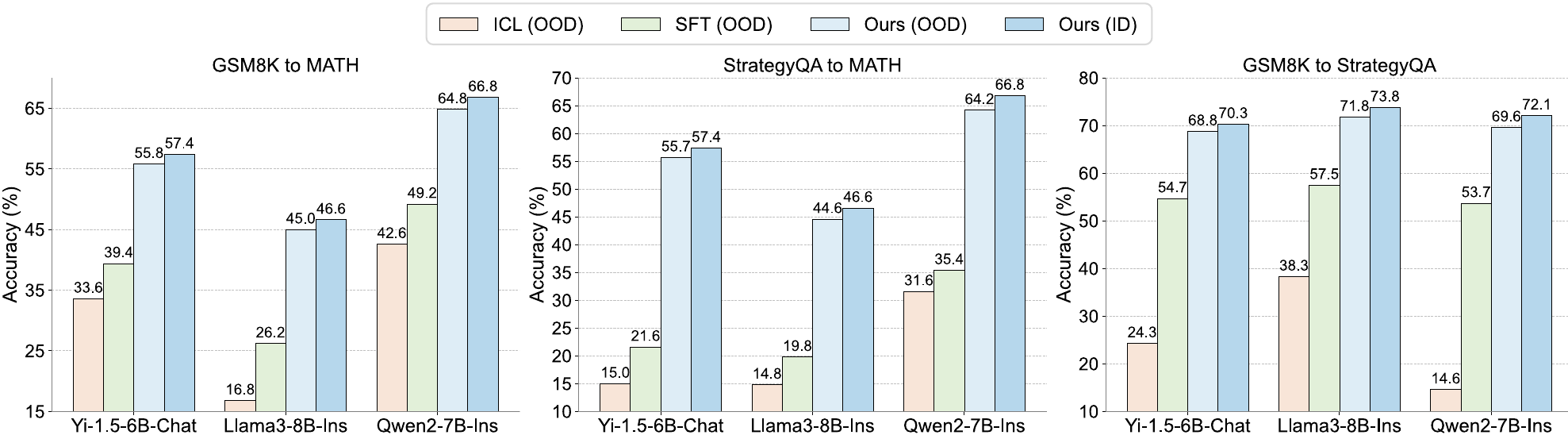}}
\caption{Cross-domain performance comparison. ``ID'' and ``OOD'' denote in-distribution and out-of-distribution, respectively. HiAR-ICL demonstrates superior generalization ability across domains compared with traditional example-based ICL methods and supervised fine-tuning (SFT) techniques.}
\label{Figure5}
\vskip -0.05 in
\end{figure*}

\textbf{Comparison with Tree-based Methods}
$\,$ We compare our method HiAR-ICL with other powerful tree-based reasoning methods in Table \ref{table2}. While existing methods like ToT and RAP face significant challenges, our method demonstrates superior performance and notable generalization across various models and datasets. The performance gap widens with increasing reasoning complexity from GSM8K to MATH. Specifically, for Llama3-8B-Instruct, our method achieves 46.6$\%$ accuracy compared to the best baseline's 42.9$\%$. Similarly for Qwen2-7B-Instruct, it improves from 53.3$\%$ to 66.8$\%$. While traditional tree search methods provide broad exploration capabilities and performance upper bound, they struggle to identify precise reasoning paths, particularly for complex reasoning tasks. Our approach addresses this limitation by leveraging prior thought cards to effectively guide reasoning based on problem attributes. Detailed results are described in Appendix Table \ref{tables30}.

\hypertarget{sec4.2}{\subsection{Computational Efficiency}\label{sec4.2}}
We further compare HiAR-ICL with rStar~\citep{qi2024mutual}, a recently proposed method that significantly expands the solution search space and achieves leading performance in tree-based reasoning. Our analysis in Figure \ref{Figure4} systematically assesses both performance and computational efficiency. HiAR-ICL achieves competitive performance with rStar while substantially reducing computational overhead. For rigor comparison, we note that the total time cost per sample includes both offline card construction time and online inference time, representing the comprehensive ``end-to-end'' time cost per sample for our approach. Detailed descriptions of time cost calculations are provided in Appendix \ref{E.6}.

Specifically, HiAR-ICL achieves substantial time reductions of 21.8$\times$ on GSM8K and 21.0$\times$ on StrategyQA through adaptive pattern selection. On the more challenging MATH dataset, our method maintains a 7.0$\times$ speedup. These efficiency gains are primarily attributed to the efficient online inference that adaptively selects appropriate reasoning patterns rather than exhaustively searching the solution space for each problem. This demonstrates HiAR-ICL's ability to balance performance and efficiency across problems of varying complexity.

\hypertarget{sec4.3}{\subsection{Out-of-Distribution Generalization}\label{sec4.3}}
Recent studies have highlighted the critical impact of distributional bias on LLMs' reliability~\citep{yuan2023revisiting,0001HH0ZWY0HGJ024}. Despite impressive in-distribution (ID) performance, these models substantially underperform when confronted with out-of-distribution (OOD) data~\citep{yang-etal-2024-unveiling}. This challenge is compounded by the inherent difficulty of acquiring sufficient training data.

In this paper, we evaluate HiAR-ICL's performance against in-context learning (ICL) and supervised fine-tuning (SFT) under OOD scenarios. To ensure a fair comparison, we use the same 200 seed samples for both thought card construction and SFT. As illustrated in Figure \ref{Figure5}, while ICL and SFT experience significant performance degradation, HiAR-ICL demonstrates remarkable resilience, preserving robust performance across multiple models and datasets. These results underscore HiAR-ICL's superior robustness and generalization capabilities, positioning it as a more reliable and adaptable solution for handling diverse reasoning tasks, including both ID and OOD data.

\begin{wraptable}{r}{0.5\linewidth}
\vspace{-12.9pt}
\centering
\caption{Weak-to-strong generalization results. `CoT' denotes direct CoT reasoning by the strong model, while `HiAR-ICL' represents our approach where weaker models (left $\to$) construct thought cards that serve as reasoning guidelines for strong model ($\to$ right) inference.}
\label{weaktostrong}
\resizebox{0.99\linewidth}{!}{
\begin{tabular}{ccccc}
\toprule
\textbf{Method} & \textbf{MATH} & \textbf{GSM8K} & \textbf{SVAMP} & \textbf{StrategyQA}\\
\midrule
\rowcolor{mygray}\multicolumn{5}{c}{\textit{Llama2-7B $\to$ GPT-4o}} \\
CoT              & 77.2 & 94.3 & 94.0 & 79.5 \\                           
HiAR-ICL  & \textbf{79.6} & \textbf{95.0} & \textbf{94.3} & \textbf{80.2} \\
\midrule
\rowcolor{mygray}\multicolumn{5}{c}{\textit{Llama2-7B $\to$ Qwen2.5-14B}} \\
CoT              & 76.2 & 94.0 & 91.0 & 69.7 \\                         
HiAR-ICL  & \textbf{80.0} & \textbf{95.8} & \textbf{92.7} & \textbf{75.7} \\  
\midrule
\rowcolor{mygray}\multicolumn{5}{c}{\textit{Llama2-7B $\to$ Qwen2.5-7B}} \\
CoT              & 76.4 & 92.0 & 92.3 & 73.2 \\                     
HiAR-ICL & \textbf{78.4} & \textbf{92.3} & \textbf{92.7} & \textbf{76.0} \\
\bottomrule
\end{tabular}}
\end{wraptable}

\hypertarget{sec4.5}{\subsection{Weak-to-Strong Generalization}\label{sec4.5}}
As described in prior work~\citep{yang-etal-2024-weak}, interactions between weak and strong models can be categorized into two primary paradigms: 1) knowledge distillation, which involves transferring capabilities or knowledge from strong models to weak models; and 2) weak-to-strong improvement, where models with limited capabilities can effectively guide the development of more advanced models. Given that knowledge distillation has been extensively studied in previous research~\citep{xu2024survey}, we focus here on HiAR-ICL's potential for weak-to-strong generalization.

In our experimental setup, we leverage a weaker model (e.g., Llama2-7B) to construct thought cards that serve as reasoning guidelines for stronger models (e.g., GPT-4o), establishing a weak-to-strong generalization framework. Table \ref{weaktostrong} presents comprehensive results across different model configurations. In weak-to-strong scenarios, our method consistently outperforms CoT baselines, achieving superior performance across all evaluated tasks. Notably, the Llama2-7B model effectively guides GPT-4o, improving performance on MATH (79.6$\%$ vs. 77.2$\%$) and GSM8K (95.0$\%$ vs. 94.3$\%$). When using Llama2-7B to guide Qwen2.5-7B and Qwen2.5-14B, we obtain significant performance improvements, especially on the challenging MATH and StrategyQA datasets. These results demonstrate that structured reasoning patterns extracted by weaker models can substantially enhance stronger models' performance, offering a promising direction for scenarios where powerful teacher models are unavailable.

\hypertarget{sec4.4}{\subsection{Ablation Studies and Analysis}\label{sec4.4}}
\textbf{Effect of Thought Cards} 
$\,$ HiAR-ICL constructs thought cards to provide high-level reasoning patterns during inference. As shown in Table \ref{table3} (Row 2), replacing structured thought cards with random action-based reasoning chains (path lengths 2-5) results in a 10.4$\%$ accuracy drop, highlighting the importance of precise prior reasoning patterns.

\textbf{Effect of Cognitive Complexity Framework}
$\,$ Given pre-constructed thought cards, if the cards are not matched based on question complexity but are instead randomly selected as guiding information, we observe a performance drop of 5.0$\%$ in Table \ref{table3} (Rows 3). This indicates that matching thought cards to the complexity of the question is crucial for optimal performance.

\begin{table*}[ht!]
\centering
\caption{Ablation results on Llama3-8B-Instruct. We report the performance after removing or replacing each component of HiAR-ICL. All modules prove essential for optimal performance.}
\vskip 0.1 in
\label{table3}
\resizebox{1.0\textwidth}{!}{
\begin{tabular}{lccccccc}
\toprule
\textbf{Model Setting} & \textbf{Module} & \textbf{MATH} & \textbf{GSM8K} & \textbf{SVAMP} & \textbf{StrategyQA} & \textbf{Average} & \textbf{$\bigtriangleup$ $(\downarrow)$}\\
\midrule
HiAR-ICL & -  & \textbf{46.6} & \textbf{89.6} & \textbf{92.7} & \textbf{73.4} & \textbf{75.6} & -\\   
\cdashline{1-8}[0.8pt/0.8pt]
\quad $-$ w/o thought cards (random actions) & \textit{Sec. \ref{3.2}}  & 33.4 & 79.9 & 84.3 & 63.3 & 65.2 & \textcolor[RGB]{34,120,5}{-10.4}\\  
\cdashline{1-8}[0.8pt/0.8pt]
\quad $-$ w/o card match (random card)  & \textit{Sec. \ref{3.3}} & 39.2 & 86.1 & 89.0 & 68.0 & 70.6 & \textcolor[RGB]{34,120,5}{-5.0}\\
\quad $-$ w/ card match (subquestion count)  & \textit{Sec. \ref{3.3}}  & 42.0 & 86.9 & 90.0 & 71.5 & 72.6 & \textcolor[RGB]{34,120,5}{-3.0}\\
\quad $-$ w/ card match (semantic)  & \textit{Sec. \ref{3.3}} & 40.2 & 86.4 & 89.0 & 70.6 & 71.5 & \textcolor[RGB]{34,120,5}{-4.1}\\
\cdashline{1-8}[0.8pt/0.8pt]
\quad $-$ w/o verification (random selection)  & \textit{Sec. \ref{3.3}}    & 40.2 & 83.3 & 90.0 & 69.6 & 70.8 & \textcolor[RGB]{34,120,5}{-4.8}\\
\quad $-$ w/ verification (SC)   & \textit{Sec. \ref{3.3}}        & 42.6 & 87.3 & 92.0 & 70.3 & 73.1 & \textcolor[RGB]{34,120,5}{-2.5}\\
\quad $-$ w/ verification (PRM)  & \textit{Sec. \ref{3.3}}        & 43.8 & 88.8 & 91.7 & 73.4 & 74.4 & \textcolor[RGB]{34,120,5}{-1.2}\\
\bottomrule
\end{tabular}
}
\end{table*}

\begin{figure*}[htp!]
\centerline{\includegraphics[width=\linewidth]{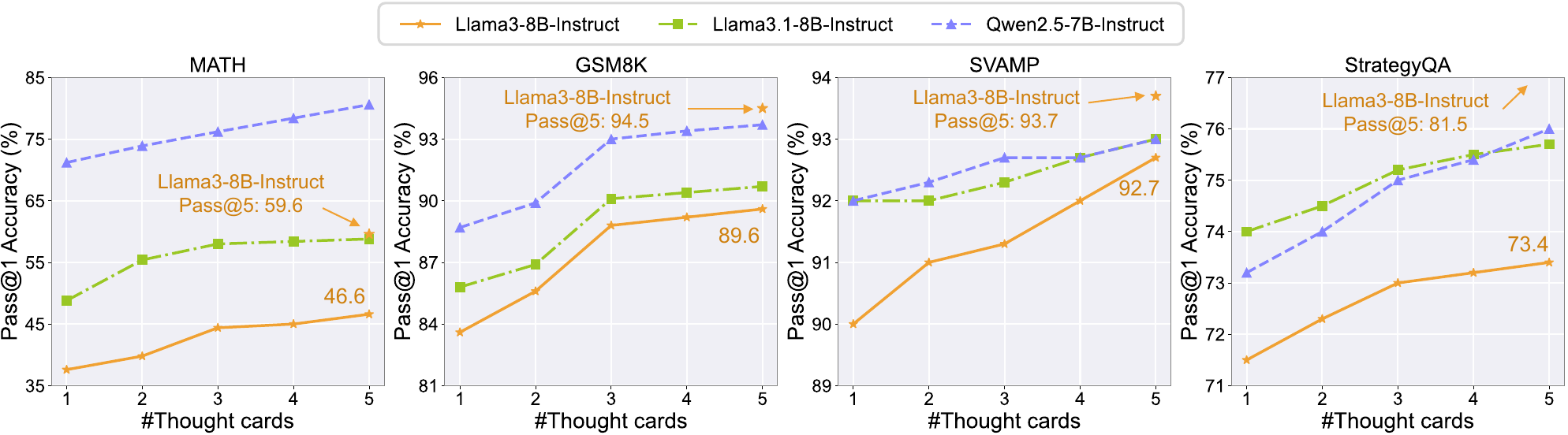}}
\vskip 0.1 in
\caption{Test-time scaling law performance. We examine the variation in HiAR-ICL performance with the number of selected reasoning guidance-providing thought cards in Section \ref{3.3}.}
\label{FigureS4}
\vskip -0.2in
\end{figure*}

\textbf{Effect of Verification Module}
$\,$ We evaluate four verification configurations: random solution selection, Self-Consistency (SC), Process Reward Model (PRM), and our combined approach (SC+PRM). Table \ref{table3} (Rows 6-8) demonstrates that path selection significantly impacts performance, with SC+PRM (our default in HiAR-ICL) achieving optimal results. While SC also performs well, even random selection shows only a 4.8$\%$ accuracy drop, indicating that our preliminary inference generates high-quality candidate solutions and confirming the effectiveness of earlier modules.

\textbf{Test-time Scaling Law in Reasoning Tasks}
$\,$ To investigate whether our method follows test-time scaling laws, we gradually increase the number of selected thought cards during inference. As shown in Figure \ref{FigureS4}, progressively adding more reasoning guidance-providing thought cards consistently improves performance for our HiAR-ICL approach. However, it is important to note that this increase in thought cards also brings about additional computational overhead. Therefore, finding an optimal balance between performance and cost becomes a key direction for future research and optimization.

\begin{wraptable}{r}{0.5\linewidth}
\vspace{-13.5pt}
\centering
\caption{Plug-and-Play validation results on Qwen2.5-1.5B. HiAR-ICL combined with GRPO-trained models further enhances performance.}
% \vskip 0.1in
\label{table50}
\resizebox{0.95\linewidth}{!}{
\begin{tabular}{llcc}
\toprule
\textbf{Method} & \textbf{MATH} & \textbf{GSM8K} & \textbf{Avg.}\\
\midrule
Zero-shot & 21.4 & 54.5 & 38.0 \\
GRPO & 43.8 & 71.9 & 57.9 \\
\cellcolor{mygray}GRPO+HiAR-ICL & \cellcolor{mygray}\textbf{52.8} & \cellcolor{mygray}\textbf{79.5} & \cellcolor{mygray}\textbf{66.2}\\
\bottomrule
\end{tabular}}
\vspace{-5pt}
\end{wraptable}

\textbf{Plug-and-Play Capability}
$\,$ Similar to ICL methods like CoT, HiAR-ICL operates as a training-free test-time inference framework compatible with post-training techniques. We demonstrated this by applying HiAR-ICL to models that have undergone GRPO training on the MATH training set. Table \ref{table50} shows that our framework consistently enhances performance when integrated with these approaches. This synergy suggests HiAR-ICL captures reasoning patterns complementary to those acquired during post-training, affirming its plug-and-play versatility. Our comprehensive results in Tables \ref{table1}, \ref{table50}, and \ref{table s12} further demonstrate HiAR-ICL's broad applicability across various model architectures and training paradigms, including base, instruction-tuned, and reinforcement learning-optimized models. This also demonstrates the generalizability and versatility of high-level thought patterns in complex reasoning tasks. Future work could explore deeper integration with post-training methods to maximize complementary benefits~\citep{zuo2025ttrl,wu2025thought}.

\textbf{Other Analysis}
$\,$ Finally, we include some additional results and analysis in the Appendix, including more performance comparisons (\ref{E.1}), combination with large reasoning models (\ref{E.18}), multi-task system (\ref{E.12}), integration with SFT (\ref{E.8}), performance across varying problem complexity (\ref{E.9}), detailed computational cost (\ref{E.6}), further ablation (\ref{E.7}), and case studies (\ref{G}).

\section{Related Work}\label{sec2}
\textbf{In-Context Learning via Examples}
\; In-context learning enables LLMs to implicitly learn reasoning strategies from few demonstrations without parameter updates~\citep{zhou-etal-2024-mystery,dong-etal-2024-survey}. Existing approaches like CoT~\citep{wei2022chain} guide step-by-step reasoning through instructions like ``Let's think step by step'' with several high-quality examples, while self-consistency~\citep{wang2023selfconsistency} improves performance by generating and aggregating multiple reasoning paths. However, these methods predominantly rely on example-level analogical learning, which exhibits limited generalization capabilities and often necessitates expert intervention for complex reasoning tasks~\citep{wang-etal-2023-label,zhao-etal-2024-unveiling,yang2024buffer,yang2025supercorrect}. In contrast, our approach shifts the focus from specific examples to high-level reasoning patterns, enabling automated, adaptive, and efficient inference across diverse complex reasoning tasks without human intervention. Notably, this paradigm synergizes LLMs' implicit reasoning capabilities with explicit external guidelines, facilitating strong performance even with relatively compact models under 10B parameters.

\textbf{Tree-based Search}
\; LLMs exhibit impressive capabilities but face challenges in complex reasoning tasks~\citep{zhao2023survey}. Tree search algorithms like MCTS~\citep{chaslot2008monte} have emerged as powerful tools to expand search spaces and enhance reasoning capabilities~\citep{koh2024tree,zhou2024language}. Recent approaches such as Tree of Thought~\citep{NEURIPS2023_271db992} and Graph of Thought~\citep{besta2024graph} explore non-linear reasoning paths through multiple LLM queries. While advanced methods like AlphaMath~\citep{chen2024alphamath} and rStar~\citep{qi2024mutual} advance reasoning capabilities, they often incur significant computational overhead. In contrast, our approach introduces a novel paradigm that strategically frontloads computational resources and incorporate prior reasoning patterns, achieving competitive performance with significant higher efficiency.

\vspace{0.2em}
\section{Conclusion}
We propose HiAR-ICL, a novel MCTS-powered reasoning paradigm that enhances in-context learning by integrating abstract reasoning patterns. Unlike conventional approaches that rely solely on example-based learning, HiAR-ICL leverages adaptive, explicit, and structured reasoning strategies, enabling LLMs to move beyond imitation toward developing genuine reasoning capabilities. This paradigm demonstrates significant performance improvements on complex reasoning tasks across multiple domains, highlighting its potential to advance efficient reasoning in LLMs. This work provides detailed empirical foundations for future research on scalable and efficient reasoning frameworks in LLMs, and opens promising directions for investigating how high-level problem-solving reasoning patterns can be automatically abstracted and applied across broader tasks and domains.

\bibliography{neurips_2025}
\bibliographystyle{ieeetr}

%%%%%%%%%%%%%%%%%%%%%%%%%%%%%%%%%%%%%%%%%%%%%%%%%%%%%%%%%%%%
\clearpage
\appendix
\section*{Appendix of HiAR-ICL}
% \parttoc
This supplementary material provides in-depth insights into our HiAR-ICL method, covering additional descriptions, experimental details, and results. The appendix is organized as follows:

\vspace{2mm}
\tableofcontents
\vspace{1.5mm}

\hrule
\vspace{3mm}

\begin{center}
    \renewcommand{\labelitemi}{}
    \begin{itemize}
        \item \textbf{\ref{tldr}. Further Discussions on HiAR-ICL and In-Context Learning}
        \item \textbf{\ref{B}. Preliminaries}
        \begin{itemize}
            \item \ref{B.1}. Overall Notations  
            \item \ref{B.2}. LLM Reasoning
            \item \ref{B.3}. In-Context Learning
        \end{itemize}
        \item \textbf{\ref{C}. Algorithm Details}
        \begin{itemize}
            \item \ref{C.1}. Action Spaces
            \item \ref{C.2}. Monte Carlo Tree Search
            \item \ref{C.3}. Thought Card Selection
            \item \ref{C.4}. Verification
        \end{itemize}
        \item \textbf{\ref{D}. More Experimental Details}
        \begin{itemize}
            \item \ref{D.1}. Models
            \item \ref{D.2}. Datasets
            \item \ref{D.3}. Baselines
            \item \ref{D.10}. Card Distribution
            \item \ref{D.5}. Evaluation Details
        \end{itemize}
        \item \textbf{\ref{E}. Supplementary Results}
        \begin{itemize}
            \item \ref{E.1}. Detailed Comparison with Powerful LLMs and Methods
            \item \ref{E.18}. Combination with Large Reasoning Models
            \item \ref{E.12}. Multi-Task System
            \item \ref{E.8}. Integration with SFT
            \item \ref{E.9}. Performance across Complexity
            \item \ref{E.6}. Detailed Computational Cost
            \item \ref{E.7}. Ablation Studies
        \end{itemize}
        \item \textbf{\ref{G}. Case Study}
    \end{itemize}
\end{center}
\vspace{5mm}
\hrule
\vspace{3mm}

\newpage
\section{Further Discussions on HiAR-ICL and In-Context Learning}\label{tldr}
In this work, we introduce HiAR-ICL, a novel paradigm within the broader in-context learning (ICL) framework. Originally proposed by prior work~\citep{NEURIPS2020_1457c0d6}, ICL refers to \textit{a process where the model is provided with natural language instructions and/or task examples and is expected to complete new task instances by predicting subsequent content}. Our proposed HiAR-ICL paradigm maintains alignment with this foundational definition while extending its scope through two main dimensions:

\textbf{Paradigm Adherence}: Although HiAR-ICL introduces a distinct pre-computation phase through MCTS to systematically construct high-level reasoning patterns (termed ``thought cards''), the paradigm fundamentally remains within the ICL framework. During inference, HiAR-ICL leverages these generated reasoning patterns explicitly formulated as natural language instructions. This approach retains the core mechanism of guiding the model via linguistic context, thus adhering strictly to the traditional conceptualization of ICL despite the augmented preparatory step.

\textbf{Context Redefinition}: We significantly expand the conventional interpretation of ``context'' within ICL. Traditionally, ``context'' refers strictly to task-specific demonstration examples. In contrast, HiAR-ICL shifts the focus from concrete example-driven contexts to more abstract, generalized reasoning patterns. This redefinition allows context to transcend task-specific boundaries, enabling models to generalize reasoning strategies effectively across diverse domains and tasks. Consequently, our broader definition enhances cross-task adaptability, significantly enriching the concept of context originally outlined by prior work~\citep{NEURIPS2020_1457c0d6}.

By redefining context from example-specific instances to higher-level reasoning abstractions, HiAR-ICL provides a robust, generalizable foundation for enhancing LLM performance, especially on complex reasoning tasks. This paradigm extension paves the way for future research that explores richer, more flexible contexts, facilitating stronger generalization capabilities across various problem-solving scenarios.

\section{Preliminaries}\label{B}
\subsection{Overall Notations}\label{B.1}
The definitions for notations are in Table \ref{table s1}.

\begin{table*}[ht!]
\caption{Notation Table in this paper.}
\vskip 0.1in
\label{table s1}
\centering
\begin{tabular}{@{\extracolsep{\fill}}cc}
\toprule
\textbf{Character} & \textbf{Meaning} \\
\midrule
$\pi_{\theta}$ & policy LLM \\
$\tau _{D}$    & specific task \\
$D$            & demonstration examples of $\tau _{D}$ in in-context learning\\
$x$            & input question / problem \\      
$y_{p}$        & predicted / decoded answer \\     
$y_{g}$        & gold standard answer \\   
$traj$         & trajectory / solution\\
$T$            & number of reasoning steps\\
$s_{t}$        & t-th reasoning step of trajectory $traj$\\
$S_{t}$        & t-th state, which consists of input x and preceding reasoning steps $(s_{1},s_{2}, ..., s_{t-1})$ \\
$a_{t}$        & t-th action based on the previous state $S_{t-1}$\\
$s$            & node $s$ in the tree structure\\
$p$            & parent node of $s$\\
$Q(s)$         & reward value of node $s$\\
$p_{\varphi } $& process reward model\\
$o_{\psi }  $  & outcome reward model\\
$D_{s}$        & seed data\\
$D_{t}$        & test data\\
\bottomrule
\end{tabular}
\end{table*}

\subsection{LLM Reasoning}\label{B.2}
LLMs have shown impressive performance across various reasoning tasks~\citep{zhao2023survey}, including mathematical~\citep{ahn-etal-2024-large} and commonsense reasoning~\citep{geva2021did}. Appropriate reasoning methods can substantially enhance LLM problem-solving capabilities, potentially transforming small models into powerful problem-solvers~\citep{fu2023specializing,qi2024mutual}. Given a policy model, $\pi_{\theta}$ (an autoregressive pre-trained LLM) and an input problem $x$, $\pi$ can autogressive generate an output sequence $output=(s_{0}, s_{1}, s_{2}, ..., s_{T})$ by predicting the next token, where $s_{0}:=x$ and $ans_{p} = s_{T}$. Each output sequence $(s_{0}, s_{1}, s_{2}, ..., s_{T})$ is termed a reasoning trajectory $traj$. The conditional probability distribution of generating the complete reasoning trajectory is:
\begin{equation}
    \pi(traj \mid x)=\prod_{t=1}^{T} \pi\left(s_{t} \mid x, s_{<t}\right)
\end{equation}

Following prior works~\citep{hao-etal-2023-reasoning, qi2024mutual}, we can conceptualize LLMs as world models, with the complex reasoning process formulated as a Markov decision process. Specifically, when addressing complex reasoning challenges in real-world scenarios, at each time step $t$, the model receives a state $S_{t-1}$, comprising the original input problem $x$ and preceding reasoning steps $(s_{0},s_{1},s_{2}, ..., s_{t-1})$. The policy model $\pi_{\theta}$ then generates the current action $a_t=\pi_{\theta}(\Phi(S_{t-1}))$, which prompts the LLM to produce the next reasoning step $s_t$. The entire process, from the initial step $s_{0}$ to the final output $s_{T}$, naturally forms a complete trajectory or chain of thought.

Recent research has focused on developing diverse methods to enhance LLMs' reasoning capabilities, including zero-shot prompting, few-shot prompting, chain-of-thought (CoT), tree-of-thought (ToT), and Monte Carlo tree search (MCTS). These approaches aim to improve overall performance through the following formulation:

\begin{equation}
P_{\pi}\left(y_p=y_g \mid Q\right) = \mathbb{E}_{\left(s_{0}, s_{1}, \cdots, s_{T}\right) \sim P_{\pi}(traj \mid x)} \left[P\left(y_p=y_g \mid s_{0}, s_{1}, \cdots, s_{T}, x\right)\right]
\end{equation}
where $P\left(y_p=y_g \mid s_{0}, s_{1}, \cdots, s_{T}, x\right)$ represents the probability of obtaining an accurate final answer given the problem $x$ and reasoning trajectory $traj$.

\subsection{In-Context Learning}\label{B.3}
Originally introduced by previous study~\citep{NEURIPS2020_1457c0d6}, in-context learning (ICL) can be described as ``\textit{A process where the model is provided with natural language instructions and/or a few examples of a task, and is then expected to complete additional instances of that task simply by predicting what should come next.}'' For a task $\tau_ {D}$, it typically involves two key components: a demonstration example space $D$ and a joint probability distribution $P(X,Y)$~\citep{zhou-etal-2024-mystery}. The task demonstration $ D={(x_{i},y_{i})}_{i=1}^{n}$ contains $n$ example pairs sampled from the joint distribution. These pairs typically consist of a problem $x$ and the corresponding solution trajectory $traj$. Therefore, the ICL-based reasoning process can be formally expressed as:
\begin{equation}
\begin{split}
    D &\sim P(X,Y), \\
    y_{p} &= \pi_{\theta}(D, x)
\end{split}
\end{equation}

Extensive research has focused on constructing high-quality examples and enriching the demonstration space $D$~\citep{luo2024incontext,zhou-etal-2024-mystery}. For example, Chain-of-Thought (CoT) reasoning~\citep{wei2022chain, kojima2022large} incorporates prompts like "Let's think step by step" alongside step-by-step reasoning examples, allowing models to emulate human-like reasoning and achieve success in complex problem-solving~\citep{sprague2024cot}. Self-Consistency~\citep{wang2023selfconsistency} further enhances performance by generating multiple reasoning paths and selecting the most consistent answer. Prompt Space~\citep{shi-etal-2024-prompt} optimizes prompt engineering for better example-based analogy. Many-Shot ICL~\citep{agarwal2024manyshot} enhances performance by scaling sample quantity, which requires substantial context-processing capabilities and computational overhead. Increasing sample volume may even degrade performance by overwhelming model capacity~\citep{liu-etal-2024-lost}.

However, this traditional ICL paradigm primarily emphasizes example-level analogical learning, with performance constrained by the selection of demonstrations. This typically necessitates human expert intervention for complex reasoning tasks~\citep{wang-etal-2023-label, zhao-etal-2024-unveiling}. In contrast, our approach shifts the focus from specific examples to high-level reasoning patterns, expanding the concept of context. This shift improves generalization, enabling fully automated, efficient inference without human intervention, even for models with fewer than 10 billion parameters. The most relevant prior work is Buffer-of-Thought (BoT)~\citep{yang2024buffer}, which designs thought templates for tasks by retrieving relevant templates to prompt LLMs. However, BoT has several limitations: (1) it relies heavily on large models like GPT-4 for template generation, (2) exhibits limited generalization to novel tasks, and (3) lacks structured reasoning mechanisms. Our approach addresses these limitations by (1) automatically constructing reasoning paths through Monte Carlo Tree Search, (2) adaptively selecting paths based on problem complexity, and (3) integrating structured reasoning processes. Notably, our framework operates entirely with smaller models while maintaining robust performance, offering both enhanced generalization and efficiency.

\section{Algorithm Details}\label{C}
\textbf{Overview}
$\;$ As shown in Figure \ref{Figure2}, HiAR-ICL consists of two components: 1) MCTS-powered thought card construction; and 2) adaptive reasoning pattern and verification.

\subsection{Action Space}\label{C.1}
Emerging research suggests that the upper bound of model reasoning capabilities is closely correlated with the available action space~\citep{wu2024comparative}. As illustrated in~\citep{qi2024mutual}, existing approaches typically involve a restricted action space (Table \ref{table s2}), which may hinder LLM's full reasoning potential. Therefore, we propose a more expansive framework, which contains five atomic reasoning actions as follows:

\begin{itemize}[leftmargin=1.5em]
    \item $(a_{1})$ \textit{System Analysis (SA)}: Analyzing the overall structure of the problem and identifying the constraints and conditions before addressing it, thereby clarifying task requirements effectively.
    \item $(a_{2})$ \textit{One-Step Thought (OST)}: Generating the next one-step thought based on the given question and the preceding reasoning steps.
    \item $(a_{3})$ \textit{Chain-of-Thought (CoT)}: Facilitating step-by-step reasoning by constructing a logical sequence of intermediate thoughts, where each step incrementally builds on the previous ones.
    \item $(a_{4})$ \textit{Divide and Conquer (DC)}: Breaking down a complex reasoning problem into several smaller subproblems and progressively solving them to achieve the overall solution.
    \item $(a_{5})$ \textit{Self-Reflection and Refinement (SRR)}: Engaging in timely reflection of prior solutions and implementing necessary refinement during the reasoning process to ensure accuracy.
\end{itemize}

\begin{table*}[htbp!]
\caption{Comparison with other tree-based search methods. Note that, most methods contain limited space. In contrast, we define a rich set of reasoning actions, thus enhancing the upper bound of model reasoning capabilities.}
\vskip 0.15 in
\label{table s2}
\centering
\begin{adjustbox}{width=0.75\textwidth}
\begin{tabular}{cc}
\toprule
\textbf{Method} & \textbf{Action Space}\\
\midrule
Tree-of-Thought~\citep{NEURIPS2023_271db992}     & \multirow{5}{*}{$a_{2}$: one-step thought} \\
AlphaMath~\citep{chen2024alphamath}          &  \\
AlphaLLM~\citep{tian2024toward}           &  \\
% MindStar~\citep{kang2024mindstar}           &   \\
ReST-MCTS*~\citep{zhang2024rest}        &  \\
\midrule
RAP~\citep{hao-etal-2023-reasoning}                 & $a_{4}$: divide and conquer        \\
BEATS~\citep{sun2024beats}                 & $a_{1}$: system analysis, $a_{2}$: one-step thought, $a_{5}$: self-refinement       \\
\midrule
MCTSr~\citep{zhang2024accessing}              & \multirow{2}{*}{$a_{3}$: chain-of-thought, $a_{5}$: self refinement} \\
LLaMA-Berry~\citep{zhang2024llama} & \\
\midrule
\textbf{Ours}    & $a_{1}$, $a_{2}$, $a_{3}$, $a_{4}$, $a_{5}$   \\
\bottomrule
\end{tabular}
\end{adjustbox}
\vskip 0.1in
\end{table*}

\begin{figure*}[htbp!]
\vskip 0.15in
\begin{center}
\centerline{\includegraphics[width=\textwidth]{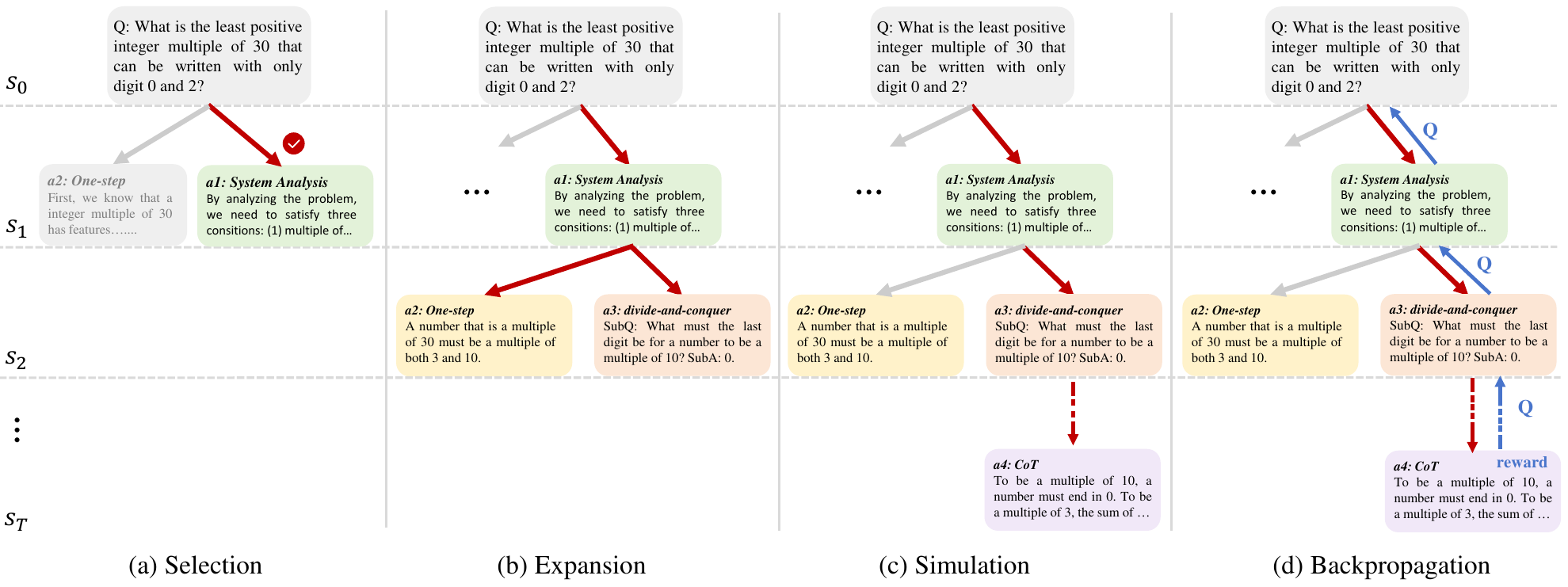}}
\caption{An illustration of four phases in an iteration of MCTS for complex reasoning tasks.}
\label{FigureS1}
\end{center}
\vskip -0.1in
\end{figure*}

\subsection{Monte Carlo Tree Search}\label{C.2}
As a heuristic search algorithm, MCTS has demonstrated remarkable success in complex reasoning and decision-making environments~\citep{chaslot2008monte,NEURIPS2021_d5eca8dc,zhou2024language}. The algorithm conceptualizes search spaces as tree structures and has achieved significant breakthroughs across various domains, most notably in game-playing AI such as AlphaGo and AlphaZero~\citep{dong2020deep}. The basic MCTS algorithm involves an iterative search process with four key steps: selection, expansion, simulation, and backpropagation. As an example in mathematical reasoning, Figure \ref{FigureS1} illustrates the four phases in an iteration, expanding the tree and then updating reward values.

Leveraging MCTS, recent approaches like rStar~\citep{qi2024mutual} exploit LLMs' intrinsic capabilities for iterative exploration to enhance complex reasoning. However, these methods employ a uniform search process across problems of varying difficulty and types, which often results in time-intensive computations. In contrast, our approach employs MCTS only during the generation of prior reasoning patterns (referred to as ``thought cards'' in Sec. \ref{3.2}) and references these thought cards during inference to achieve efficient reasoning. This design enables HiAR-ICL to adaptively match reasoning strategies to the complexity of each problem, significantly reducing the time complexity compared to traditional tree search methods. Additionally, it maintains a comprehensive search space and overall performance, thus achieving an optimal trade-off between efficiency and effectiveness.

For reward values in MCTS, we avoid introducing an additional reward model for scoring for simplicity. Given the current skepticism regarding the self-rewarding capabilities of LLMs, alternative methods are necessary. Inspired by the principle that actions leading to correct answers should be rewarded more frequently, we aim to increase their likelihood of selection in future MCTS tree expansions. Following prior work~\citep{qi2024mutual}, we define the reward value as the likelihood (or confidence) of self-consistency via majority voting. Note that this principle applies only to leaf nodes. The Q-values of intermediate nodes are initially set to 0 and are dynamically updated during the backpropagation process, as described in Equation~\ref{eq1}.

\subsection{Thought Card Selection}\label{C.3}
Within our proposed cognitive complexity framework, we define three metrics: (1) Subquestion Count (SC): quantifying the number of decomposable subproblems; (2) Problem Condition Complexity (PCC): measuring the number of distinctive problem conditions; and (3) Semantic Similarity (SS): assessing the semantic distance between the target problem and the seed dataset. As presented in Table \ref{table3} in \textit{Section \ref{sec4.4}}, our empirical analysis reveals minimal statistical variations among these metrics. Consequently, we adopt PCC as the primary metric. We conjecture that future research could productively explore sophisticated integration strategies, such as weighted ranking summation, to comprehensively leverage these three metrics and develop a more nuanced quantification of reasoning problem complexity.

\subsection{Verification}\label{C.4}
In this paper, we introduce a simple yet effective two-stage verification method. First, we apply process-supervision scoring to evaluate each reasoning path. The top-3 highest-scoring paths then undergo self-consistency checks to determine the final solution. Our experiments confirm that even these straightforward self-consistency checks effectively identify precise reasoning chains without additional supervision. In practical implementation, we utilize off-the-shelf pre-trained models: Llama3.1-8B-ORM-Mistral-Data\footnote{\href{https://huggingface.co/RLHFlow/Llama3.1-8B-ORM-Mistral-Data}{RLHFlow/Llama3.1-8B-ORM-Mistral-Data}} (ORM) and Llama3.1-8B-PRM-Mistral-Data\footnote{\href{https://huggingface.co/RLHFlow/Llama3.1-8B-PRM-Mistral-Data}{RLHFlow/Llama3.1-8B-PRM-Mistral-Data}} (PRM). By default, our experimental results are reported based on PRM verification. Moreover, as demonstrated in Table \ref{table3} in \textit{Section \ref{sec4.4}}, our method still achieves notable performance even in scenarios lacking a readily available verification model, such as when relying solely on the simple yet effective method, self-consistency. In future work, we aim to explore more sophisticated verification approaches to more precisely select the optimal reasoning trajectory.

\section{More Experimental Details}\label{D}
\subsection{Models}\label{D.1}
HiAR-ICL is a general approach applicable to various LLMs. In our experiments, we evaluate its effectiveness using powerful open-source models: Llama3-8B-Instruct~\citep{dubey2024llama}, Llama3.1-8B-Instruct~\citep{meta_llama_2023}, Yi-1.5-6B-Chat~\citep{young2024yi}, Qwen2-7B-Instruct~\citep{yang2024qwen2}, and Qwen2.5-7B/14B-Instruct~\citep{qwen25}. Unless stated otherwise, all models are the Instruct version. In \textit{\ref{E.8} Integration with SFT}, we also analyze the performance of HiAR-ICL combined with base models. By focusing on LLMs with parameter counts generally under 10B, we aim to demonstrate the robustness and efficiency of our method. We expect that applying HiAR-ICL to small language models will achieve results comparable to or exceeding closed-source LLMs.

\begin{table*}[htbp!]
\caption{Detailed information on the datasets utilized in the main result. Seed datasets are used to construct thought cards, and test sets for final evaluation.}
\vskip 0.1in
\label{table s3}
\centering
\begin{adjustbox}{width=0.8\textwidth}
\begin{tabular}{llcc}
\toprule
\textbf{Category} & \textbf{Dataset} & \textbf{\#Seed Samples} & \textbf{\#Test Samples} \\
\midrule
\multirow{2}{*}{Arithmetic}      
& GSM8K~\citep{cobbe2021training}     & 200 & 1319 \\
& SVAMP~\citep{Patel2021AreNM}        & 200 & 300  \\
\midrule
\multirow{2}{*}{Mathematics}  
& MATH~\citep{hendrycks2021measuring} & 200 & 500  \\
& AMC~\citep{li2024numinamath} & 200 (MATH training set) & 40  \\
\midrule
Commonsense     
& StrategyQA~\citep{geva2021did}      & 200 & 687  \\
\midrule
Science     
& GPQA$_{Diamond}$~\citep{rein2024gpqa} & 200 (MATH training set) & 198  \\
\bottomrule
\end{tabular}
\end{adjustbox}
\end{table*}

\subsection{Datasets}\label{D.2}
The datasets utilized in this paper are listed in Table \ref{table s3}.

\begin{itemize}[leftmargin=1.36em]
    \item \textbf{GSM8K~\citep{cobbe2021training}}: This dataset contains 7,473 training and 1,319 testing grade-school math word problems, which require between 2 to 8 steps for resolution. The solutions primarily involve performing a sequence of basic arithmetic operations (addition, subtraction, multiplication, and division) to arrive at the final answer.
    \item \textbf{SVAMP~\citep{Patel2021AreNM}}: Similar to GSM8K, this dataset contains a 1,000-sample testing set of elementary-level mathematical word problems (MWPs). These problems exhibit diverse structural variations and challenge large language models (LLMs) to generate precise numerical values or equations for solution. Following the existing dataset partition\footnote{\href{https://huggingface.co/datasets/ChilleD/SVAMP}{https://huggingface.co/datasets/ChilleD/SVAMP}}, we extract 200 samples from the 700-sample training set as the seed dataset and utilize the remaining 300-sample test set for evaluation.
    \item \textbf{MATH~\citep{hendrycks2021measuring}}: This dataset includes 12,500 challenging competition-level mathematics problems. Each problem is accompanied by a complete step-by-step solution, which can be leveraged to teach models to generate answer derivations and explanations. Following prior work~\citep{qi2024mutual,zhang2024rest}, we utilize MATH500 as our test set for a fair comparison, a representative and highly challenging 10$\%$ subset of MATH.
    \item \textbf{AMC~\citep{li2024numinamath}}: This dataset contains problems from the American Mathematics Competitions, specifically focusing on high-school level mathematical challenges. The AMC problems typically require sophisticated mathematical reasoning and creative problem-solving strategies beyond basic arithmetic. For our evaluation, we use a subset of 40 carefully selected problems that demonstrate diverse mathematical concepts and varying complexity levels.
    \item \textbf{StrategyQA~\citep{geva2021did}}: This dataset comprises 2,780 examples, each consisting of a strategy question, its decomposition, and supporting evidence paragraphs. We utilize the 687 examples from the test set for evaluation. This dataset challenges models to perform multi-hop reasoning across diverse domains of commonsense knowledge.
    \item \textbf{GPQA~\citep{rein2024gpqa}}: GPQA (Graduate-level Physics Questions and Answers) is a PhD-level scientific knowledge reasoning benchmark that contains complex questions requiring advanced domain expertise. The Diamond198 subset we use for evaluation consists of 198 particularly challenging questions that test deep scientific knowledge and sophisticated reasoning capabilities. These questions often require integration of multiple scientific concepts and principles to reach the correct solution.
\end{itemize}
For the seed dataset $D_{s}$, we randomly sample 200 instances from the training sets of each dataset to construct thought cards. The sampling's randomness ensures algorithmic robustness. Additionally, only 200 samples are used for tree search, and cross-distribution transfer capabilities have been validated in the main text. Therefore, in practical scenarios, publicly available datasets can be utilized to construct the seed dataset, enabling the construction of thought cards and final reasoning. This ensures the generality of the approach.

\subsection{Baselines}\label{D.3}
For baselines compared in this paper, we conduct experiments on ICL approaches such as CoT and self-consistency to record their performance. For other tree-based methods and closed-source models, we directly reference their original reported results when accessible. In cases where reported results are unavailable but open-source code is provided, we reproduce experiments following the official official settings.

\subsection{Card Distribution}\label{D.10}
We analyze the length and diversity of action chains across datasets of varying difficulty. As shown in Table~\ref{tab:action_chain_stats}, both metrics increase with problem complexity. However, many reasoning patterns, such as ``decompose and think step by step,'' remain consistent across datasets. This consistency helps explain the strong out-of-domain generalization of thought cards, as demonstrated in Figure~\ref{Figure5}. Additionally, not all five actions are required for each problem. Simpler problems, like Q1 in Figure~\ref{Figure2}, may require only one action ($CoT$), whereas more complex problems, like Q4, involve longer chains ($SA \to OST \to SRR\to CoT$). The number and complexity of actions thus depend heavily on the specific problem at hand.

\begin{table}[htbp]
    \centering
    \caption{Statistics of reasoning complexity and action chains across datasets.}
    \vskip 0.1in
    \begin{tabular}{lccc}
        \toprule
        \textbf{Dataset} & \textbf{Difficulty} & \textbf{Avg. Length of Action Chain} & \textbf{Number of Action Chains} \\
        \midrule
        GSM8K & $\star$ & 3.09 & 35 \\
        MATH & $\star\star$ & 3.59 & 42 \\
        AMC & $\star\star\star$ & 3.67 & 45 \\
        \bottomrule
    \end{tabular}
    \label{tab:action_chain_stats}
\end{table}

\subsection{Evaluation Details}\label{D.5}
As detailed in the main paper, we evaluate our approach using two primary metrics. First, we report accuracy as our primary evaluation metric, where correctness is determined by comparing the model's final answer with the ground truth. To ensure consistent answer extraction, we require the LLM to explicitly state its solution following a predefined format (e.g., ``The answer is''). Second, we measure the average reasoning time to assess our method's computational efficiency relative to existing search-based approaches. All time measurements were recorded on the same device (A100 GPUs), thereby ensuring fair comparisons.

\begin{table*}[ht!]
\centering
\caption{Comparison with leading LLMs. The best results in each box are highlighted in \textbf{bold}. Results for all baseline models are sourced from corresponding official websites when accessible. HiAR-ICL (Qwen2.5-7B-Instruct) outperforms larger models and powerful closed-source models.}
\vskip 0.1 in
\label{table s5}
\begin{adjustbox}{width=0.7\textwidth}
\begin{tabular}{lrccc}
\toprule
\textbf{Model} & \textbf{Size} & \textbf{MATH} & \textbf{GSM8K} & \textbf{Average} \\
\midrule
\rowcolor{mygray}\multicolumn{5}{c}{\textit{\bfseries Closed-Source Models}} \\
GPT-4o-0806~\citep{gpt4o}               & -    & \textbf{77.2} & 96.1 & \textbf{86.6} \\
GPT-4o mini-0718~\citep{gpt4o}          & -    & 70.2 & 93.2 & 81.7 \\
Claude-3.5-Sonnet-0620~\citep{claude}   & -    & 71.1 & \textbf{96.4} & 83.8 \\
Claude-3-Opus-0304~\citep{claude3}      & -    & 60.1 & 95.0 & 77.6 \\
Gemini-1.5-Pro~\citep{gemini}           & -    & 67.7 & 90.8 & 79.3 \\
GPT-4~\citep{achiam2023gpt}             & -    & 64.5 & 94.2 & 79.4 \\
GPT-3.5~\citep{gpt35}                   & -    & 43.1 & 81.6 & 62.4 \\
\midrule
\rowcolor{mygray}\multicolumn{5}{c}{\textbf{\textit{Open-Source Models ($>$30B)}}} \\
Llama3.1-405B-Instruct~\citep{meta_llama_2023}      & 405B & 73.8 & \textbf{96.8} & \textbf{85.3} \\
Nemotron4-340B-Instruct~\citep{adler2024nemotron}   & 340B & 41.1 & 92.3 & 66.7 \\
% DeepSeek-V2.5~\citep{liu2024deepseek}               & 236B & 74.7 & 95.1 & 84.9 \\
Mixtral-8x22B-Instruct~\citep{mixtral822}           & 141B & 54.1 & 88.2 & 71.2 \\
Mixtral-large2-Instruct~\citep{mixtrallarge2}       & 123B & 69.9 & 92.7 & 81.3 \\
Qwen2-72B-Instruct~\citep{yang2024qwen2}            & 72B  & 69.0 & 93.2 & 81.1 \\
% NuminaMath-72B CoT~\citep{numina_math_7b}           & 72B  & 66.7 & 90.8 & 78.8 \\
Llama3.3-70B-Instruct~\citep{meta_llama_2023}       & 70B  & \textbf{77.0} & - & 77.0 \\
Llama3.1-70B-Instruct~\citep{meta_llama_2023}       & 70B  & 68.0 & 95.1 & 81.6 \\
Llama3-70B-Instruct~\citep{dubey2024llama}          & 70B  & 50.4 & 93.0 & 71.7 \\
Yi-1.5-34B-Chat~\citep{young2024yi}                 & 34B  & 50.1 & 90.2 & 70.2 \\
\midrule
\rowcolor{mygray}\multicolumn{5}{c}{\textbf{\textit{Ours ($\le$14B)}}} \\
Qwen2.5-14B-instruct~\citep{qwen25}                 & 14B  & \textbf{81.4} & \textbf{95.8} & \textbf{88.6} \\
Qwen2.5-7B-instruct~\citep{qwen25}                  & 7B   & 80.6 & 93.7 & 87.2 \\
Qwen2-7B-instruct~\citep{yang2024qwen2}             & 7B   & 66.8 & 91.8 & 79.3 \\
Yi-1.5-6B-Chat~\citep{young2024yi}                  & 6B   & 57.4 & 86.4 & 71.9 \\
Llama3-8B-Instruct~\citep{dubey2024llama}           & 8B   & 46.6 & 89.6 & 68.1 \\
Llama3.1-8B-Instruct~\citep{meta_llama_2023}        & 8B   & 58.0 & 90.7 & 74.4 \\
\bottomrule
\end{tabular}
\end{adjustbox}
\end{table*}

\begin{table*}[ht!]
\centering
\caption{Comparision of HiAR-ICL's reasoning performance with tree-based methods and advanced ICL methods across four benchmarks. ``Reinforced ICL'' refers to the advanced ICL technique from Many-Shot In-Context Learning~\citep{agarwal2024manyshot}, evaluated with 10-shot examples. ``CoT+AS'' denotes directly providing a high-level action sequence in the instruction. ToT and ReST-MCTS* serve as representative baselines from tree-based methods. The best results in each box are highlighted in \textbf{bold}. All models are instruct versions.}
\label{tables30}
\vskip 0.1 in
\begin{adjustbox}{width=0.98\linewidth}
\begin{tabular}{clccccccl}
\toprule
\multirow{2}{*}{\textbf{Model}}  & \multirow{2}{*}{\textbf{Method}} & \multicolumn{1}{c}{\textbf{Mathematics}} & \multicolumn{2}{c}{\textbf{Arithmetic}} &  \textbf{Commonsense} & \multirow{2}{*}{\textbf{Average}} \\
\cmidrule(lr){3-3} \cmidrule(lr){4-5} \cmidrule(lr){6-6} 
 &  & MATH  & GSM8K & SVAMP  & StrategyQA & \\
\midrule
\multirow{5}{*}{GPT-4o~\citep{gpt4o}} 
                                      & CoT+AS & 80.8 & 95.5 & 94.0 & 80.8   & 87.8 \\
                                      & Reinforced ICL & 81.0 & 95.5 & 94.0 & 77.7 & 87.0 \\
                                      & ToT         & 75.8 & 95.2 & 92.7 & 75.5 & 84.8 \\
                                      & ReST-MCTS*  & 78.0 & 94.6 & 93.3 & 76.9 & 85.7 \\
                                      & \cellcolor{mygray}Ours    & \cellcolor{mygray}\textbf{84.8} &  \cellcolor{mygray}\textbf{96.0} & \cellcolor{mygray}\textbf{94.7} & 
                                      \cellcolor{mygray}\textbf{82.2} &  \cellcolor{mygray}\textbf{89.4} \\
\midrule
\multirow{5}{*}{Qwen2.5-14B~\citep{qwen25}} 
                                      & CoT+AS & 78.2 & 94.3 & 91.7 & 72.5   & 84.2 \\
                                      & Reinforced ICL & 78.8 & 93.7 & 93.7 & 70.3 & 84.1 \\
                                      & ToT         & 71.4 & 93.8 & 92.0 & 73.3 & 82.6 \\
                                      & ReST-MCTS*  & 74.1 & 94.0 & 92.3 & 71.4 & 82.9 \\
                                      & \cellcolor{mygray}Ours    & \cellcolor{mygray}\textbf{81.4} &  \cellcolor{mygray}\textbf{95.8} & \cellcolor{mygray}\textbf{93.7}  & 
                                      \cellcolor{mygray}\textbf{77.3}  &  \cellcolor{mygray}\textbf{87.0} \\
\midrule
\multirow{5}{*}{Qwen2.5-7B~\citep{qwen25}}  
                                      & CoT+AS & 77.2 & 92.7 & 92.7 & 71.8   &83.6 \\
                                      & Reinforced ICL & 76.6 & 91.3 & 92.0 & 72.5 & 83.1 \\
                                      & ToT         & 68.4 & 91.7 & 92.1 & 71.3 & 80.8 \\
                                      & ReST-MCTS*  & 72.2& 92.4 & 91.7 & 69.5 & 81.5 \\
                                      & \cellcolor{mygray}Ours    & \cellcolor{mygray}\textbf{80.6} &  \cellcolor{mygray}\textbf{93.7} & \cellcolor{mygray}\textbf{93.0}  & 
                                      \cellcolor{mygray}\textbf{76.0} &  \cellcolor{mygray}\textbf{85.9} \\
\midrule
\multirow{5}{*}{Qwen2-7B~\citep{yang2024qwen2}}    
                                      & CoT+AS & 58.2 & 87.7 & 89.7 & 68.5   &76.0 \\
                                      & Reinforced ICL & 54.8 & 87.9 & 91.7 & 66.8 & 75.3 \\
                                      & ToT         & 53.3 & 79.0 & 85.7 & 66.7 & 71.1 \\
                                      & ReST-MCTS*  & 52.4& 82.3 & 86.8 & 64.9 & 71.6\\                          
                                      & \cellcolor{mygray}Ours    & \cellcolor{mygray}\textbf{66.8} &  \cellcolor{mygray}\textbf{91.8} & \cellcolor{mygray}\textbf{92.7}  & 
                                      \cellcolor{mygray}\textbf{72.0}  &  \cellcolor{mygray}\textbf{80.8} \\
\midrule
\multirow{5}{*}{Yi-1.5-6B~\citep{young2024yi}}       
                                      & CoT+AS & 47.8 & 81.0& 88.3 & 66.5   &70.9 \\
                                      & Reinforced ICL & 40.4 & 78.4 & 88.3 & 62.3 & 67.4 \\
                                      & ToT         & 43.8 & 76.1 & 81.0 & 66.7 & 66.9 \\
                                      & ReST-MCTS*  & 40.6& 78.6 & 84.7 & 59.8 & 65.9\\  
                                      & \cellcolor{mygray}Ours    & \cellcolor{mygray}\textbf{57.4}  & \cellcolor{mygray}\textbf{86.4} & \cellcolor{mygray}\textbf{91.3}  & 
                                      \cellcolor{mygray}\textbf{70.3}  &  \cellcolor{mygray}\textbf{76.4} \\
\midrule
\multirow{5}{*}{Llama3-8B~\citep{dubey2024llama}}  
                                      & CoT+AS & 32.6 & 83.0& 90.3 & 68.1   &68.5 \\
                                      & Reinforced ICL & 25.6 & 81.0 & 89.0 & 65.2 & 65.2 \\
                                      & ToT         & 13.6 & 69.0 & 79.8 & 60.4 & 55.7 \\
                                      & ReST-MCTS*  & 34.2& 75.5 & 88.0 & 65.0 & 65.7\\  
                                      & \cellcolor{mygray}Ours    & \cellcolor{mygray}\textbf{46.6}  & \cellcolor{mygray}\textbf{89.6} & \cellcolor{mygray}\textbf{92.7}  &  \cellcolor{mygray}\textbf{73.4} &  \cellcolor{mygray}\textbf{75.6} \\
\bottomrule
\end{tabular}
\end{adjustbox}
\vskip -0.1in
\end{table*}

\section{Supplementary Results}\label{E}
This section presents supplementary results and analysis, including: comparison with powerful LLMs and methods (\ref{E.1}), combination with large reasoning models (\ref{E.18}), multi-task system (\ref{E.12}), integration with SFT (\ref{E.8}), performance across problem complexity (\ref{E.9}), detailed computational cost (\ref{E.6}), and ablation studies (\ref{E.7}, effect of atomic action).

\subsection{Detailed Comparison with Powerful LLMs and Methods}\label{E.1} 
Table \ref{table s5} presents a performance comparison between our method and leading open-source and closed-source models. By employing HiAR-ICL, both our 14B and 7B Qwen2.5 models achieve superior results, surpassing many powerful models with over 100B parameters. Notably, on the challenging MATH dataset, our method demonstrates a significant performance advantage, underscoring its effectiveness in complex reasoning tasks. This highlights the strength of our approach in achieving robust reasoning capabilities while utilizing relatively small models.

We have extended Table \ref{table1} to advanced ICL methods (e.g. reinforced ICL), and tree-based methods (e.g. ToT) for a more comprehensive comparison in Table \ref{tables30}. HiAR-ICL consistently outperforms advanced ICL methods and tree-based methods. For example, Llama3-8B’s accuracy on MATH improved from 17.8\% (ToT) to 46.6\% (HiAR-ICL), a 2.6$\times$ improvement.

\subsection{Combination with Large Reasoning Models}\label{E.18} 
As shown below, HiAR-ICL enables smaller models (7B) to approach the performance of larger models like GPT-4o, highlighting its potential to enhance reasoning in small models. It can also be combined with larger reasoning models like QwQ-32B-Preview for even better performance.
\begin{table}[htbp]
    \centering
    \caption{Performance of HiAR-ICL combined with large reasoning models on the MATH dataset.}
    \vskip 0.1in
    \begin{tabular}{lc}
        \toprule
        \textbf{Method} & \textbf{MATH Acc (\%)} \\
        \midrule
        QwQ-32B-Preview & 90.6 \\
        o1-preview & 85.5 \\
        \rowcolor{mygray}
        HiAR-ICL (Qwen2.5-7B) & 80.6 \\
        \rowcolor{mygray}
        HiAR-ICL (QwQ-32B-Preview) & \textbf{93.6} \\
        \bottomrule
    \end{tabular}
    \label{tab:math_results}
\end{table}

\subsection{Multi-Task System}\label{E.12}
Multi-task learning frameworks~\citep{caruana1997multitask} represent a promising approach to enhancing general system performance. Here, we aim to explore a generalized system capable of simultaneously handling multiple tasks.

Specifically, we extract 50 exemplars from each of three reasoning domains: mathematical, arithmetic, and commonsense reasoning. These 150 seed data samples are used to construct high-level thought cards, providing guidance signals for subsequent inference. As shown in Table \ref{tables20}, HiAR-ICL still achieves competitive performance, further demonstrating the versatility of our reasoning paradigm. This result may provide new insights for future research on multi-task general systems, suggesting that exploring high-level patterns across tasks could represent a promising direction.

\begin{table}[htp!]
\caption{Multi-task performance of Llama3-8B-Instruct across various reasoning tasks. `MT' refers to generating mixed thought cards to guide subsequent inference, while `ICL' represents the best performance with few-shot CoT. We observe that Our approach is also well-suited as a general multi-task reasoning system.}
\label{tables20}
\vskip 0.1in
\centering
\adjustbox{max width=0.75\linewidth}{
\begin{tabular}{@{\extracolsep{\fill}}ccccccc}
\toprule
\textbf{Method} & \textbf{MATH} & \textbf{GSM8K} & \textbf{SVAMP} & \textbf{StrategyQA} & \textbf{Average}\\
\midrule
ICL           & 17.8 & 74.5 & 81.0 & 68.4 & 60.4\\
HiAR-ICL      & 46.6 & 89.6 & 92.7 & 73.4 & 75.6\\
HiAR-ICL-MT   & 44.2 & 87.4 & 91.7 & 71.2 & 73.7\\
\bottomrule
\end{tabular}
}
\end{table}

\begin{table*}[ht!]
\centering
\caption{Integration results of HiAR-ICL with SFT on Llama3-8B and Llama3.1-8B base models. $\bigtriangleup$ indicates the performance gain of HiAR-ICL compared to the best baseline. The results show that HiAR-ICL not only significantly enhances the performance of base models but also continues to provide substantial improvements even after large-scale supervised fine-tuning (SFT). This demonstrates the strong compatibility of our method with SFT and its robustness in boosting reasoning performance across multiple tasks.}
\label{table s12}
\vskip 0.1in
\adjustbox{max width=1.0\textwidth}{
\begin{tabular}{llcccccc}
\toprule
\textbf{Model} & \textbf{Setting} & \textbf{MATH} & \textbf{GSM8K} & \textbf{SVAMP} & \textbf{StrategyQA} & \textbf{Average} & \textbf{$\bigtriangleup$ $(\uparrow)$}\\
\midrule
\multirow{3}{*}{Llama3-8B} & Zero-shot CoT & 5.8 & 17.7 & 30.0 & 9.6 & 15.8 & -\\   
 & Few-shot CoT & 13.2 & 39.4 & 56.7 & 61.1 & 42.6 & -\\ 
& HiAR-ICL (Ours) & \textbf{34.4} & \textbf{81.4} & \textbf{90.0} & \textbf{70.7} & \textbf{69.1} & \textcolor[RGB]{192,0,0}{+26.5}\\  
\cdashline{2-8}[0.8pt/0.8pt]

\multirow{3}{*}{Llama3-8B-Instruct} & Zero-shot CoT & 5.8 & 68.3 & 70.9 & 57.2 & 50.5 & -\\   
 & Few-shot CoT & 17.8 & 74.5 & 81.0 & 68.4 & 60.4 & -\\   
& HiAR-ICL (Ours) & \textbf{46.6} & \textbf{89.6} & \textbf{92.7} & \textbf{73.4} & \textbf{75.6} & \textcolor[RGB]{192,0,0}{+15.2}\\  
\midrule
\multirow{3}{*}{Llama3.1-8B} & Zero-shot CoT & 7.6 & 18.4 & 29.0 & 5.7 & 15.2 & -\\   
 & Few-shot CoT & 14.8 & 40.4 & 59.0 & 61.3 & 43.9 & -\\   
& HiAR-ICL (Ours) & \textbf{38.4} & \textbf{82.3} & \textbf{90.0} & \textbf{71.1} & \textbf{70.5} & \textcolor[RGB]{192,0,0}{+26.6}\\  
\cdashline{2-8}[0.8pt/0.8pt]

\multirow{3}{*}{Llama3.1-8B-Instruct} & Zero-shot CoT & 18.0 & 61.5 & 69.3 & 52.4 & 50.3 & -\\   
 & Few-shot CoT & 47.2 & 76.6 & 82.0 & 63.6 & 67.3 & -\\    
& HiAR-ICL (Ours) & \textbf{58.0} & \textbf{90.7} & \textbf{93.0} & \textbf{75.7} & \textbf{79.4} & \textcolor[RGB]{192,0,0}{+12.1}\\  
\bottomrule
\end{tabular}
}
\vskip 0.1in
\end{table*}

\subsection{Integration with SFT}\label{E.8}
Integrating our training-free paradigm, HiAR-ICL, with Supervised Fine-Tuning (SFT) presents an opportunity to expand its applicability and scalability, particularly for complex reasoning tasks. Here, we conduct exploratory experiments to assess the compatibility between the two approaches. As shown in Table \ref{table s12}, we evaluate both pretrained checkpoints and instruction-tuned versions of Llama3-8B and Llama3.1-8B.

Our results show that, when applied to base models (Llama3-8B and Llama3.1-8B), HiAR-ICL achieves substantial performance improvements through its structured guidance mechanism, significantly outperforming both zero-shot and few-shot baselines. The performance gains are particularly notable, reaching improvements of up to 26.6$\%$ on average across multiple benchmarks. More importantly, when integrated with instruction-tuned models, HiAR-ICL continues to deliver remarkable enhancements, with consistent improvements across all evaluated tasks, demonstrating strong synergy between structured reasoning and SFT techniques.

These findings suggest that our reasoning paradigm can be effectively integrated with SFT to enhance model performance. In future work, we plan to explore more sophisticated integration strategies, such as injecting structured long-chain reasoning data (i.e., reasoning patterns encoded on thought cards) into the model's inference process through SFT. This could involve techniques inspired by chain-of-preference optimization (CPO)~\citep{zhang2024chain}, further optimizing model reasoning efficiency and scalability.

\subsection{Performance across Problem Complexity}\label{E.9}
As shown in Table \ref{tables10}, we present the performance of Llama3-8B-Instruct and Qwen2.5-7B-Instruct on Zero-shot CoT, Few-shot CoT+SC, and our method, HiAR-ICL, across different difficulty levels of the challenging MATH dataset. Taking Qwen2.5-7B-Instruct as an example, our approach improves performance across all levels, with an average accuracy boost of 2.6$\%$ for the easier levels 1-3. Notably, for the more difficult level 4, the improvement reaches +7.7$\%$. This indicates that our method has the potential to solve more challenging problems and enhance reasoning performance. This may be due to the introduction of high-level reasoning patterns, which help LLMs find a clearer solution more quickly.

\begin{table}[htbp!]
\caption{Performance variations of Llama3-8B-Instruct and Qwen2.5-7B-Instruct across different difficulty levels on MATH. We list the result of zero-shot CoT, fewshot CoT+SC, and our method. The best results are highlighted in \textbf{bold}. $\bigtriangleup$ indicates the performance gain of HiAR-ICL compared to the best baseline.}
\vskip 0.1in
\centering
\label{tables10}
\begin{adjustbox}{width=0.7\linewidth}
\begin{tabular}{cccccccc}
\toprule
\textbf{Method} & \textbf{1} & \textbf{2} & \textbf{3} & \textbf{4} & \textbf{5} & \textbf{Average} & \textbf{$\bigtriangleup$ $(\uparrow)$}\\
\midrule
\rowcolor{mygray}\multicolumn{8}{c}{\textit{\bfseries Llama3-8B-Instruct}}\\
CoT      & 53.5 & 34.3 & 22.1 & 8.2 & 3.7 & 17.8 & -\\
CoT+SC   & 62.5 & 52.6 & 38.1 & 19.5 & 9.7 & 28.8 & -\\
HiAR-ICL & \textbf{88.4} & \textbf{75.6} & \textbf{56.2} & \textbf{32.0} & \textbf{16.5} & \textbf{46.6} & \textcolor[RGB]{192,0,0}{+17.8}\\
\midrule
\rowcolor{mygray}\multicolumn{8}{c}{\textit{\bfseries Qwen2.5-7B-Instruct}}\\
CoT      & 95.3 & 87.7 & 79.0 & 67.2 & 40.3 & 68.6 & -\\
CoT+SC   & 95.3 & 90.0 & 91.4 & 73.4 & 52.2 & 76.4 & -\\
HiAR-ICL & \textbf{97.7} & \textbf{94.5} & \textbf{92.3} & \textbf{81.1} & \textbf{53.7} & \textbf{80.6} & \textcolor[RGB]{192,0,0}{+4.2}\\
\bottomrule
\end{tabular}
\end{adjustbox}
\end{table}

\subsection{Detailed Computational Cost}\label{E.6}
In this section, we present a more comprehensive analysis of the time cost of HiAR-ICL. Additionally, we include an extended comparison with representative MCTS-based methods to show that HiAR-ICL's high efficiency.

For clarity, we first define the time cost. For downstream evaluation tasks, the average time cost per sample is formally defined in Equation \ref{avg_time}, where ``total offline construction time'' represents the time cost of constructing thought cards using 200 seed samples, and ``total online inference time'' represents the inference time on $n$ downstream samples. Importantly, the offline construction overhead is fixed and independent of downstream task scale. As the number of test samples ($n$) increases, the amortized offline cost per sample diminishes.
\begin{equation}
t_{\text{avg}} = \frac{\text{total offline construction time}}{n} + \frac{\text{total online inference time}}{n}
\label{avg_time}
\end{equation}
We conducted experiments on Llama3-8B on three downstream evaluation tasks: GSM8K, MATH and STG, the results are shown in Table \ref{time_costs}. HiAR-ICL maintains high efficiency while achieving strong accuracy on downstream tasks. For example, on the STG task, it achieves the highest accuracy while using only 5\% of the time required by rStar, demonstrating the efficiency of the proposed method.

\begin{table}[htp!]
\centering
\caption{Time cost comparison between HiAR-ICL and representative tree-based methods. The best results are highlighted in bold. `off' and 'on' denote average offline and online time cost per sample.}
\vskip 0.1in
\label{time_costs}
\adjustbox{width=\columnwidth}{
\begin{tabular}{ccccccc}
\toprule
\textbf{Method} & \textbf{GSM8K Acc (\%)} & \textbf{GSM8K Time Cost (↓)} & \textbf{MATH Acc (\%)} & \textbf{MATH Time Cost (↓)} & \textbf{STG Acc (\%)} & \textbf{STG Time Cost (↓)}\\
\midrule
ToT           & 69.0 & 623.5s & 13.6 & 809.8s & 60.4&567.3s\\
ReST-MCTS*	  & 75.5 & 677.8s & 34.2 & 467.2s & 65.0&503.5s\\
rStar         & \textbf{91.1} & 554.6s & 42.9 & 1105.8s & 71.5&423.8s\\
HiAR-ICL      & 89.6 & \textbf{25.5s (off: 8.2 + on: 17.3)} & \textbf{46.6} & \textbf{159.5s (off: 50.8 + on: 108.7)} & \textbf{73.4}&\textbf{20.2s (off: 6.5 + on: 13.7)}
\\
\bottomrule
\end{tabular}}
\end{table}

\subsection{Ablation Studies}\label{E.7}
In addition to the ablation studies on modules 1 (thought cards construction), and 2 (reasoning pattern selection, verification) in the main text, we present additional results on module 1 (\hyperlink{E.7.2}{\textit{effect of each atomic action}}).

\vspace{0.7em} \noindent 
\hypertarget{E.7.2}{\textbf{Effect of Each Atomic Action}}
$\,$ We conduct an ablation study to evaluate the effectiveness of each atomic action within our proposed action space using the Llama3-8B-Instruct model. Our experiments in Table \ref{table s6} highlight the critical importance of individual action components. Compared to existing approaches with typically restrictive action spaces (as shown in Table \ref{table s2}), our expanded action space yields significant performance improvements.
The study may open avenues for future research, particularly in developing adaptive action space selection mechanisms for broader problem domains. We aim to enhance approach generalization by dynamically customizing action spaces, with a specific focus on emerging multimodal reasoning tasks.

\begin{table}[htp!]
\centering
\caption{Effectiveness of each atomic action in this paper. We evaluate on Llama3-8B-Instruct.}
\vskip 0.1in
\label{table s6}
\adjustbox{max width=0.7\linewidth}{
\begin{tabular}{cccc}
\toprule
\textbf{Action Space} & \textbf{MATH} & \textbf{StrategyQA} & \textbf{Average} \\
\midrule
$a_{3}$ (i.e. CoT)      & 17.8 & 68.4 & 43.1\\ 
$a_{3}$  + $a_{4}$ & 42.6 & 69.9 & 56.3\\
$a_{2}$ + $a_{3}$ + $a_{4}$ & 43.2 & 70.3 & 56.8\\ 
$a_{1}$ + $a_{2}$ + $a_{3}$ + $a_{4}$ & 46.0 & 72.6 & 59.3\\ 
All ($a_{1}$ + $a_{2}$ + $a_{3}$ + $a_{4}$ + $a_{4}$) & 46.6 & 73.4 & 60.0\\ 
\bottomrule
\end{tabular}
}
\end{table}

\section{Case Study}\label{G}
\begin{figure*}[htbp!]
\centering
\includegraphics[width=0.92\textwidth]{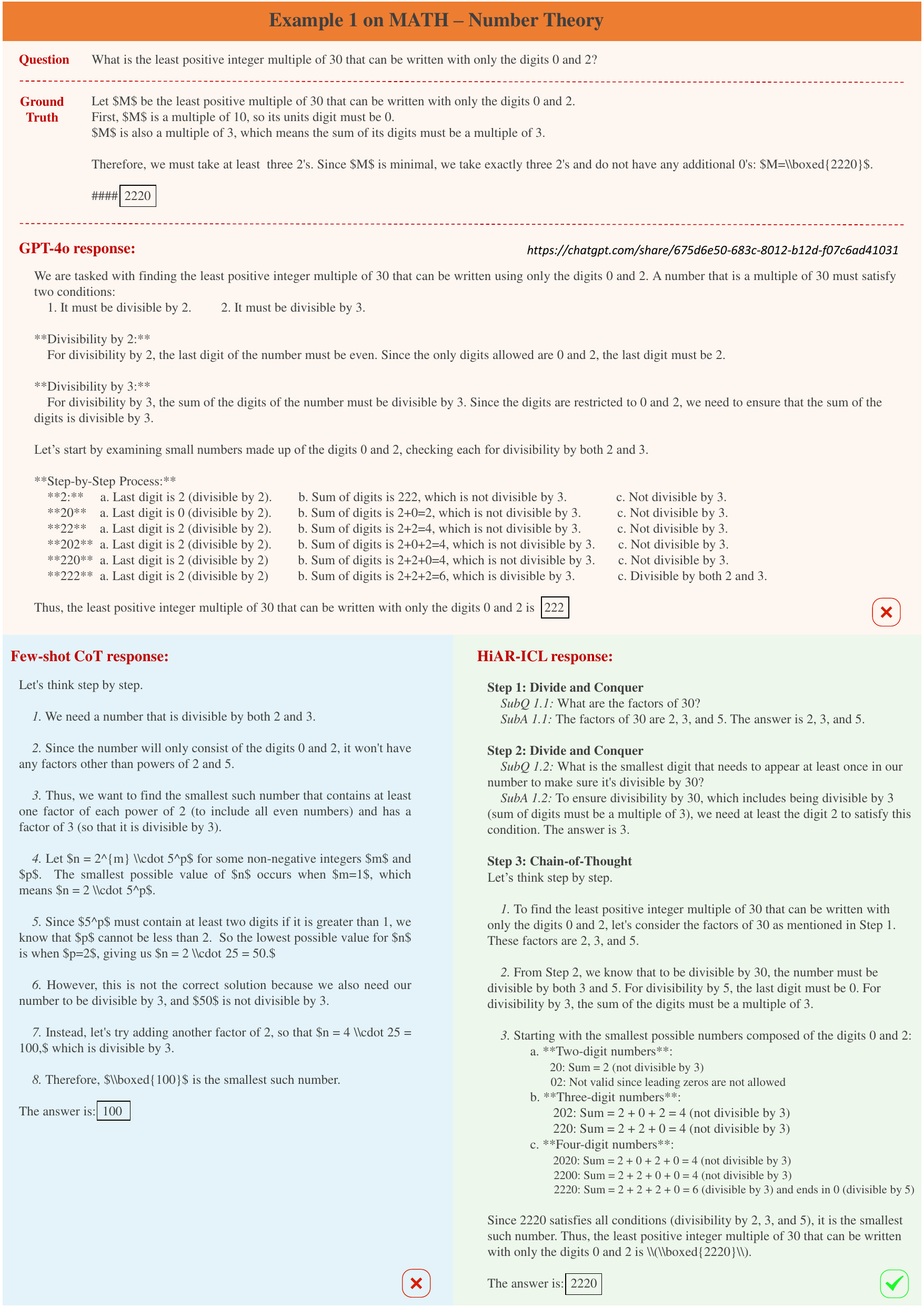}
\caption{Example 1 about Number Theory on the MATH dataset. GPT-4o results are available at \href{https://chatgpt.com/share/675d6e50-683c-8012-b12d-f07c6ad41031}{GPT-4o results}}
\label{FigureS10}
\end{figure*}

\begin{figure*}[htbp!]
\centering
\includegraphics[width=0.92\textwidth]{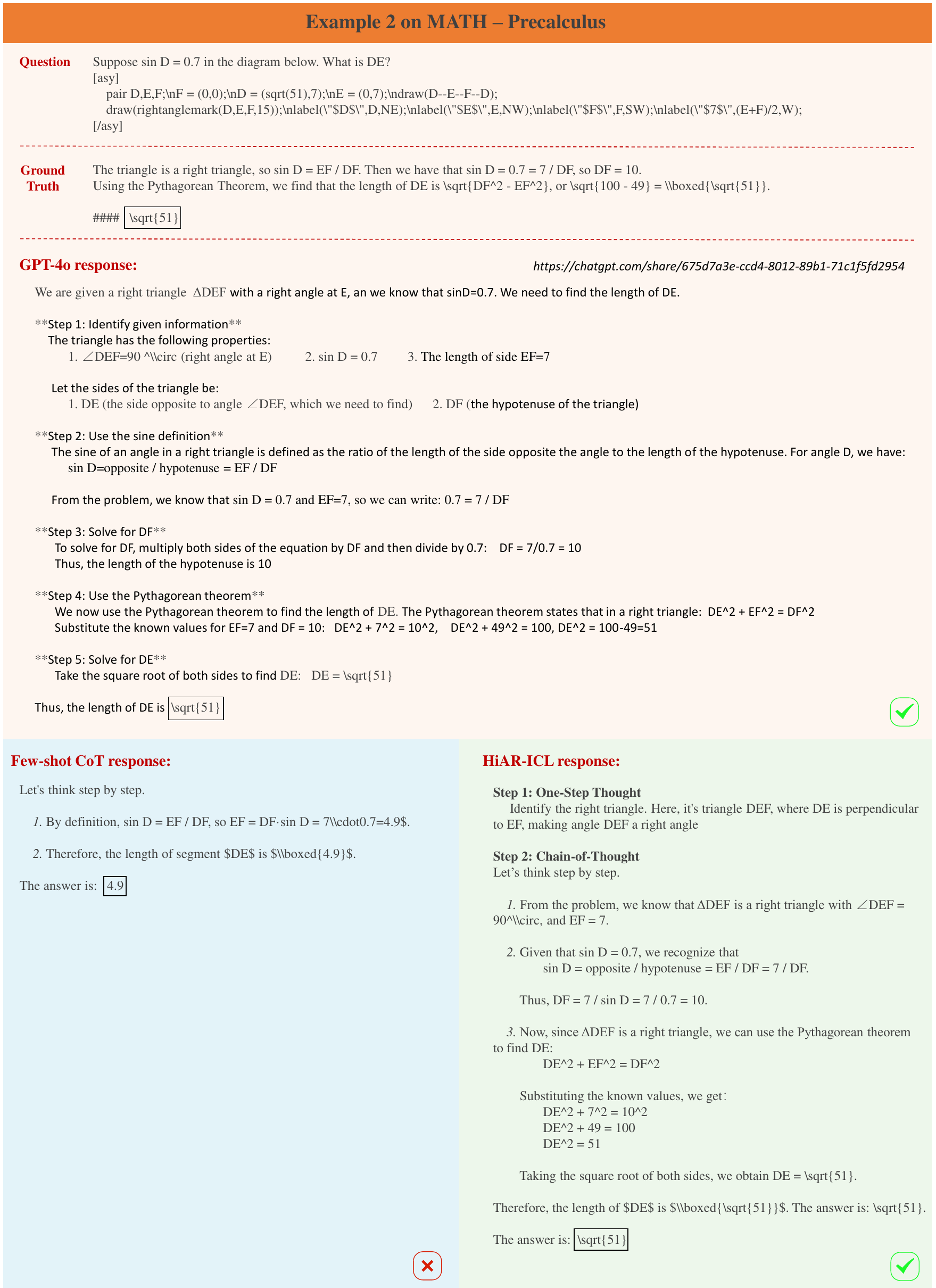}
\caption{Example 2 about Precalculus on the MATH dataset. GPT-4o results are available at \href{https://chatgpt.com/share/675d7a3e-ccd4-8012-89b1-71c1f5fd2954}{GPT-4o results}}
\label{FigureS11}
\end{figure*}

\begin{figure*}[htbp!]
\centering
\includegraphics[width=0.92\textwidth]{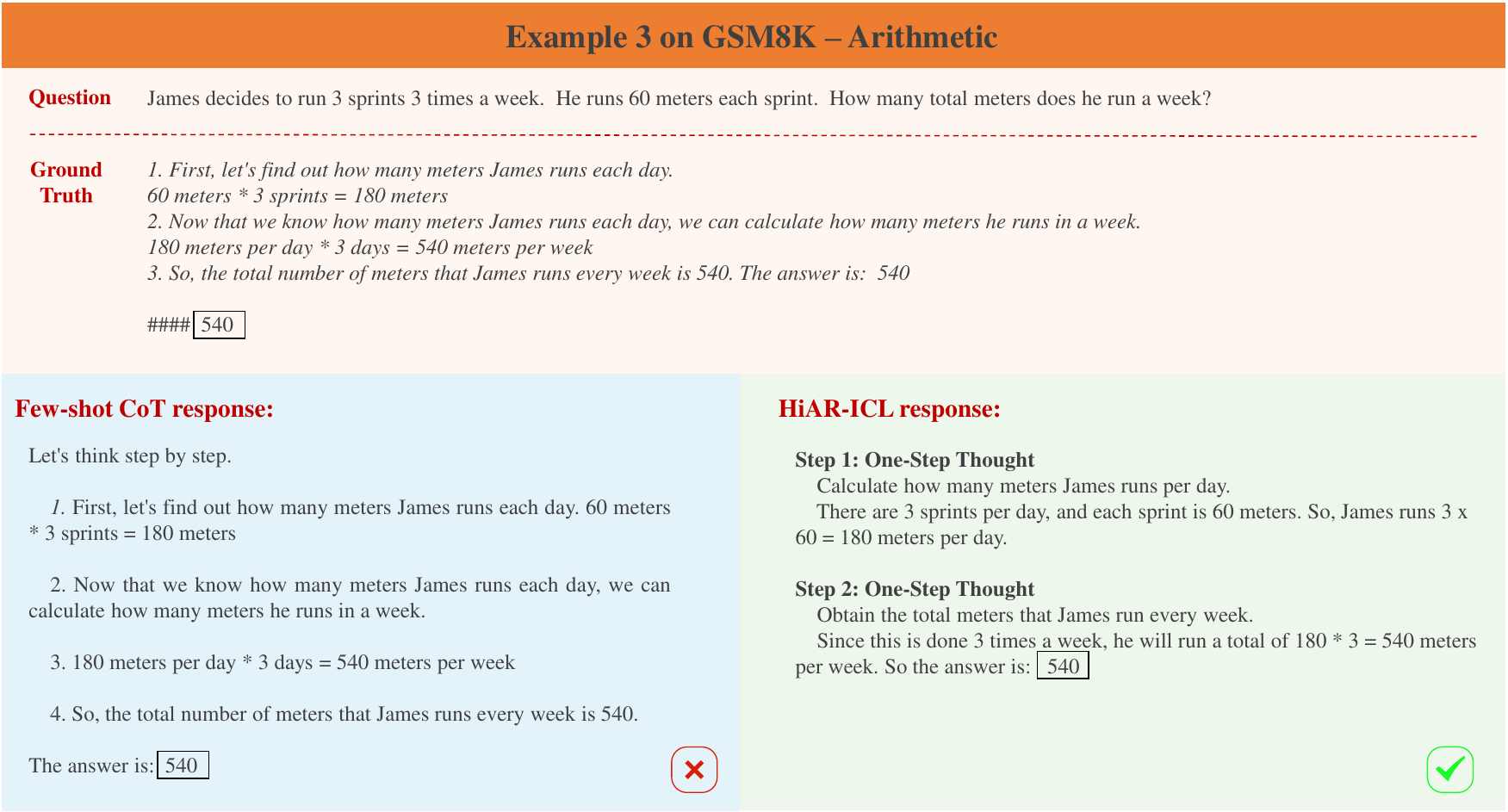}
\caption{Example 3 about arithmetic on the GSM8K dataset.}
\label{FigureS12}
\end{figure*}

\begin{figure*}[htbp!]
\centering
\includegraphics[width=0.92\textwidth]{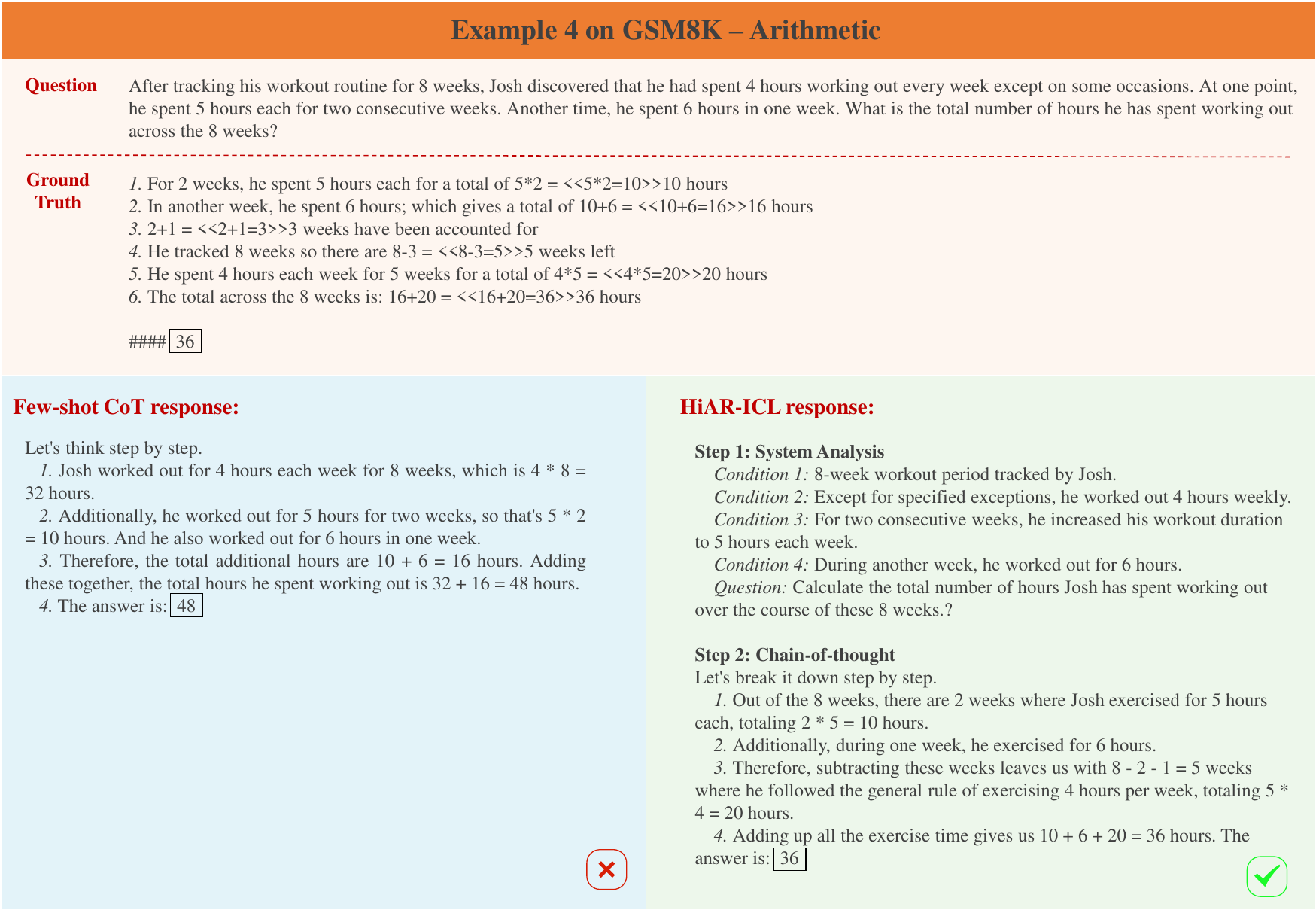}
\caption{Example 4 about arithmetic on the GSM8K dataset.}
\label{FigureS13}
\end{figure*}

\begin{figure*}[htbp!]
\centering
\includegraphics[width=0.92\textwidth]{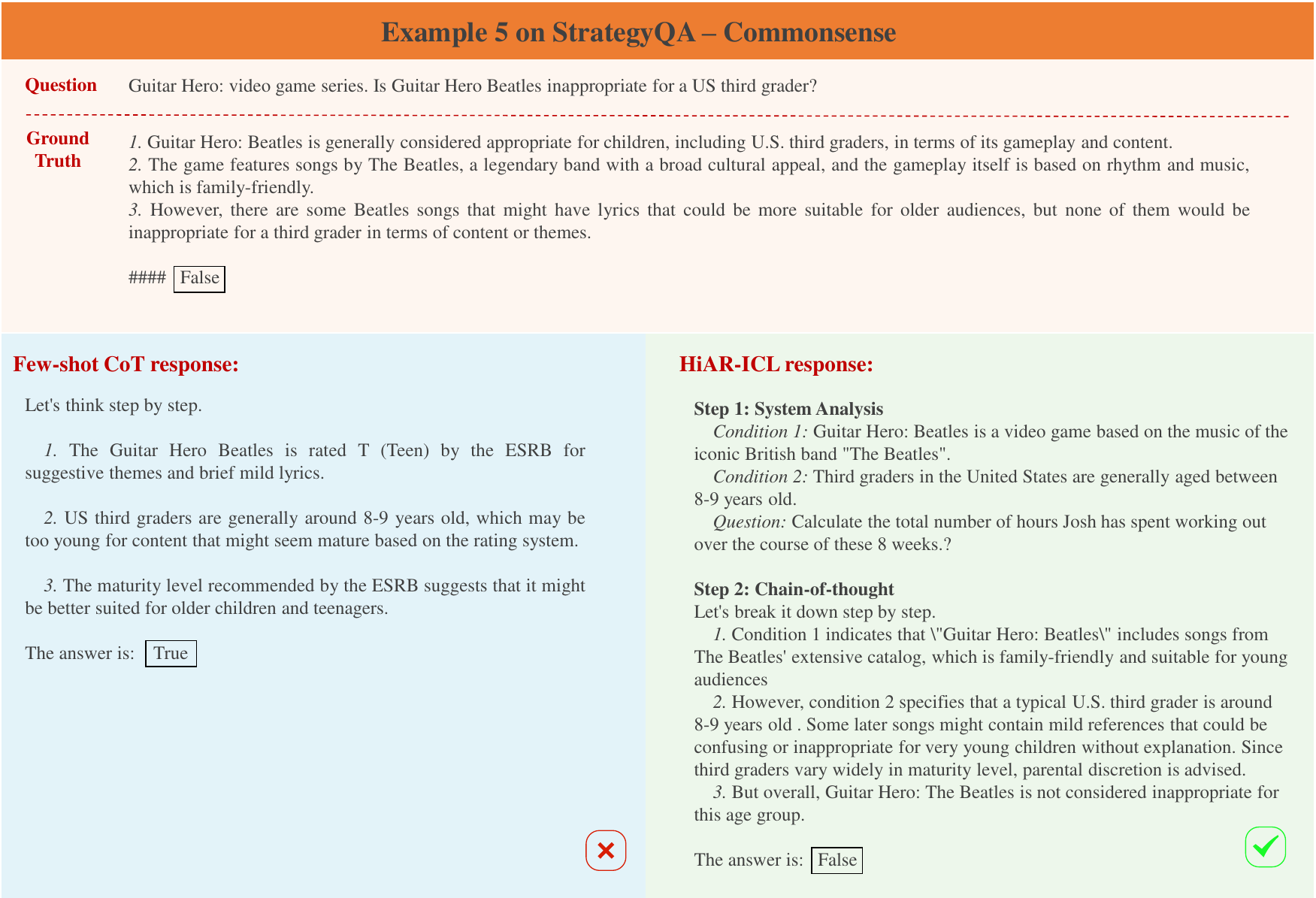}
\caption{Example 5 about commonsense on the StrategyQA dataset.}
\label{FigureS14}
\end{figure*}

Figures \ref{FigureS10}-\ref{FigureS14} present detailed problem-solving examples employing our method and in-context learning. Our approach demonstrates notable advantages in addressing relatively challenging problems, effectively leveraging the rich action space and guidance-providing thought cards to achieve precise reasoning. For instance, in Figure \ref{FigureS10}, both ICL and GPT-4o (\href{https://chatgpt.com/share/675d6e50-683c-8012-b12d-f07c6ad41031}{results}) output incorrect results, while our method with Qwen2.5-7B-Instruct provides a logically clear solution process and ultimately arrives at the correct answer.

\newpage

\end{document}